\documentclass[11pt,a4paper]{article}

\usepackage[utf8]{inputenc}
\usepackage[T1]{fontenc}
\usepackage[margin=2.5cm]{geometry}
\usepackage{mathpazo}              % Palatino serif
\usepackage[scaled=0.92]{helvet}   % Helvetica sans
\usepackage{microtype}
\usepackage{graphicx}
\usepackage{booktabs}
\usepackage{longtable}
\usepackage{array}
\usepackage{tabularx}
\usepackage{multirow}
\usepackage{enumitem}
\usepackage{xcolor}
\usepackage{natbib}
\usepackage[colorlinks=true,linkcolor=teal!70!black,citecolor=teal!70!black,urlcolor=teal!70!black]{hyperref}
\usepackage{titlesec}
\usepackage{fancyhdr}
\usepackage{textcomp}
\usepackage{amsmath}
\usepackage{amssymb}

\definecolor{accentteal}{HTML}{1A6B5C}
\definecolor{lightaccent}{HTML}{E6F0ED}
\definecolor{warmhighlight}{HTML}{FDF6F0}
\definecolor{highlightborder}{HTML}{C9875F}
\definecolor{inkgrey}{HTML}{3D4752}

\titleformat{\section}
  {\Large\bfseries\color{accentteal}}{\thesection}{1em}{}
  [\vspace{-0.5em}\textcolor{accentteal!30}{\rule{\linewidth}{0.4pt}}]
\titleformat{\subsection}
  {\large\bfseries}{\thesubsection}{0.8em}{}
\titleformat{\subsubsection}
  {\normalsize\bfseries\color{inkgrey}}{\thesubsubsection}{0.8em}{}

\newsavebox{\noteboxcontent}

\newsavebox{\warnboxcontent}

\newcommand{\srcrule}{\textcolor{accentteal!55}{\rule{\linewidth}{0.4pt}}}
\newcommand{\hl}[1]{{\setlength{\fboxsep}{1pt}\colorbox{lightaccent}{#1}}}
\newcommand{\bad}[1]{{\setlength{\fboxsep}{1pt}\colorbox{highlightborder!30}{#1}}}
\newcommand{\dn}{\vspace{2pt}$\big\downarrow$\vspace{1pt}}
\newcommand{\srccard}[3]{%
  \fcolorbox{accentteal}{white}{\parbox[t]{#1}{%
    \footnotesize\textbf{\color{accentteal}#2}\\[2pt]
    \scriptsize\ttfamily #3}}}

\newcolumntype{L}[1]{>{\raggedright\arraybackslash}p{#1}}
\newcolumntype{C}[1]{>{\centering\arraybackslash}p{#1}}

\DeclareRobustCommand{\rev}[1]{#1}
\newcommand{\revon}{}
\begin{document}

% Title page
\begin{titlepage}
\centering
\vspace*{3cm}

{\small\textsc{\color{accentteal}\textbf{Study Protocol  -  Observational Retrospective Study}}}

\vspace{1.5cm}

{\LARGE\bfseries CONTEXT-PCa: Evaluating the role of clinical context in MRI-based AI for detecting clinically significant prostate cancer \par}
% Clinical Context Determines the Development,\\ Utility and Generalisability of Artificial\\ Intelligence for Prostate MRI Diagnosis:\\[0.3em] A Retrospective Cohort Study\par
\vspace{1.5cm}

{\large
Shiqi~Huang\textsuperscript{1,2},
Weixi~Yi\textsuperscript{1,2},
Shaheer~Ullah~Saeed\textsuperscript{1--5},
Yipei~Wang\textsuperscript{1,2},
Wen~Yan\textsuperscript{1,2},
Natasha~Thorley\textsuperscript{6,7},
Shonit~Punwani\textsuperscript{6,7},
Aqua~Asif\textsuperscript{8,9},
Alexander~Ng\textsuperscript{8,9},
Veeru~Kasivisvanathan\textsuperscript{8--11},
Mark~Emberton\textsuperscript{8,10},
Yipeng~Hu\textsuperscript{1,2,*}\par}

\vspace{0.8cm}

{\small
\textsuperscript{1}UCL Hawkes Institute, University College London, London, UK\\
\textsuperscript{2}Department of Medical Physics and Biomedical Engineering, University College London, London, UK\\
\textsuperscript{3}Centre for Bioengineering, Queen Mary University of London, London, UK\\
\textsuperscript{4}School of Engineering and Materials Science, Queen Mary University of London, London, UK\\
\textsuperscript{5}Digital Environment Research Institute, Queen Mary University of London, London, UK\\
\textsuperscript{6}Centre for Medical Imaging, University College London, London, UK\\
\textsuperscript{7}Department of Radiology, University College London Hospitals NHS Foundation Trust, London, UK\\
\textsuperscript{8}Division of Surgery and Interventional Science, University College London, London, UK\\
\textsuperscript{9}Centre for Urology Imaging, Prostate, AI and Surgical Studies (COMPASS) Research Group, University College London, London, UK\\
\textsuperscript{10}Department of Urology, University College London Hospitals NHS Foundation Trust, London, UK\\
\textsuperscript{11}Department of Urology, Comprehensive Cancer Center, Medical University of Vienna, Vienna, Austria\\[0.5em]
\textsuperscript{*}Corresponding author: Yipeng Hu (\href{mailto:yipeng.hu@ucl.ac.uk}{yipeng.hu@ucl.ac.uk})\par}

\vfill

\end{titlepage}

% Table of contents
\tableofcontents
\newpage

% ==========================================================
\section{Background and Rationale}
\label{sec:background}

Prostate cancer is the most commonly diagnosed male malignancy worldwide, with 1,414,249 new cases and 375,000 deaths annually (2020 GLOBOCAN estimates)~\citep{Sung2021}. Diagnosis increasingly relies on magnetic resonance imaging (MRI) followed by targeted biopsy, a pathway supported by clinical trials (PROMIS~\citep{Ahmed2017}, PRECISION~\citep{Kasivisvanathan2018}) and prospective studies (MRI-FIRST~\citep{Rouviere2019} and 4M~\citep{vanderLeest2019}). Bi-parametric MRI (bpMRI), which omits gadolinium-based contrast enhancement in favour of T2-weighted and diffusion-weighted sequences alone, has gained traction as a faster, cheaper, and contrast-free alternative with comparable diagnostic accuracy. In the UK alone, approximately 200,000 men undergo prostate MRI each year.

The Prostate Imaging-Reporting and Data System version 2.1 (PI-RADS v2.1)~\citep{Turkbey2019} provides a structured framework for reporting prostate MRI, yet interpretation remains subjective, radiologist-dependent, and constrained by a global shortage of trained readers. Clinically significant prostate cancer (csPCa), defined following the PI-CAI challenge labelling procedure~\citep{Saha2024} as ISUP grade group $\geq$2 (Gleason score $\geq$3+4) in any core, is one of the primary targets in clinical detection pathways because it is associated with with an increased risk of adverse oncological outcomes and generally requires closer surveillance or consideration of active treatment~\citep{Epstein2016,Cornford2024}.

Clinical information available within the prostate cancer diagnostic pathway is highly heterogeneous. Established predictors such as PSA, PSA density, prostate volume, age, DRE findings, radiologist score, ethnicity, and family history are routinely used for risk assessment~\citep{Roobol2013,Ankerst2014,Turkbey2019,Cornford2024}. Other factors, including longitudinal PSA change, BMI, lifestyle behaviours, laboratory abnormalities, \rev{comorbidities,} and \rev{periprostatic adipose tissue (PPAT), are suspected to be associated with the diagnosis and prediction of prostate cancer and} may provide additional clinical context, although their predictive contribution may be more population- and setting-dependent.
% TODO: add citations for the PPAT and prostate cancer association

Despite their established clinical relevance, the added value of clinical variables within MRI-based AI systems remains uncertain. Existing studies have reported mixed results, with some showing no statistically significant benefit from adding clinical information and others demonstrating clear improvements; moreover, performance can depend substantially on how clinical and imaging information are integrated~\citep{Roest2023,Roest2024,Woznicki2020,Hiremath2021}. This leaves unresolved which clinical variables provide genuine complementary value beyond imaging, which may introduce redundant or dataset-specific associations that impair generalisability, and how heterogeneous clinical and imaging information should best be integrated within AI models.

% A broad range of clinical variables carry independent predictive value for csPCa, including prostate-specific antigen (PSA), PSA change, PSA density, prostate volume, patient age, body mass index (BMI), blood and urine test results, digital rectal examination (DRE) findings, PI-RADS score, ethnicity, smoking status, alcohol use, comorbidities (diabetes, hypertension, cardiovascular disease, COPD, asthma, chronic kidney disease), family history of cancer (including prostate cancer specifically), and history of other malignancies. However, the role of these clinical variables within AI-based diagnostic systems for prostate MRI is incompletely characterised. Most existing deep learning studies either use imaging alone or incorporate only a limited set of clinical covariates, leaving it unclear which clinical features contribute meaningful added value and how different integration strategies affect model performance.

% \begin{warnbox}
% \textbf{Knowledge gap:} There is a lack of systematic investigation into the role of clinical variables within AI systems for prostate MRI-based csPCa prediction - specifically, which variables carry incremental value, how they should be integrated with imaging features, and how the breadth and quality of clinical context influence model performance and generalisability.
% \end{warnbox}

% ==========================================================
\section{Objectives}
\label{sec:objectives}

This study investigates the role of clinical variables - their breadth, source, and quality - in the development, utility, and generalisability of an AI system for bi-parametric MRI (bpMRI)-based diagnosis of clinically significant prostate cancer (csPCa, ISUP $\geq$2). The objectives below give effect to that question.

% We therefore systematically evaluate a broad spectrum of routinely available clinical variables, both individually and jointly, across multiple image–clinical integration strategies and clinically relevant patient subgroups.

\subsection{Primary Objective}

To determine whether combining clinical variables with bpMRI improves the detection of csPCa relative to bpMRI alone, evaluated on the held-out UCLH test set. Clinical-only models are evaluated alongside as a reference for the performance attainable without imaging.

\subsection{Secondary Objectives}

\begin{enumerate}[label=\arabic*., leftmargin=2em]
  \item Quantify the incremental predictive value of individual clinical variables through add-one-in and leave-one-out ablation analyses.
  \item Determine whether the value of clinical context depends on the stage at which it is combined with imaging features - early, mid, or late combining.
  \item Determine whether a clinical variable's contribution depends on its source and quality: structured \rev{extraction, automated image analysis, or} extraction from free \rev{text;} its missingness \rev{rate; and, for free-text variables,} its validated extraction reliability.
  \item Characterise performance heterogeneity across subgroups defined by age, ethnicity, PSA level, and radiologist score.
  \item Identify which clinical variables drive individual predictions, and whether their contributions are consistent across combining strategies.
  \item \rev{Evaluate} the image-only \rev{and image-context models} on \rev{external cohorts from primary diagnostic} and \rev{population screening settings.}
\end{enumerate}

% ==========================================================
\section{Study Design}
\label{sec:design}

This is a retrospective cohort study using routinely collected electronic health record (EHR) data and clinically acquired MRI studies from University College London Hospitals (UCLH) NHS Foundation Trust. The study follows a diagnostic prediction model development and internal validation design, reported in accordance with the STROBE statement~\citep{vonElm2007} for observational studies and the TRIPOD+AI guideline~\citep{Collins2024} for prediction model studies involving AI/machine learning.

\subsection{Setting and Study Period}

Data are drawn from the UCLH Epic EHR system, covering the period from April 2019 (system go-live) to March 2024. UCLH is a tertiary academic centre in central London with a high-volume prostate MRI service.

Clinical prostate MRI is performed across several UCLH sites: the UCLH Main Building, where three scanners are potentially used for prostate MRI; Westmoreland Street Hospital; the Macmillan Cancer Centre; the Grafton Way Building; and the Royal National ENT and Eastman Dental Hospitals. Approximately 12 consultant radiologists reported prostate MRI over the study period. The cohort therefore spans multiple sites, scanners, and reporting radiologists.

Prostate MRI at UCLH is reported using the Likert scale, which records the reporting radiologist's overall level of suspicion for clinically significant cancer on a subjective five-point scale. Unlike a PI-RADS category, which is assigned from the images alone, a Likert score permits the radiologist to weigh clinical context such as PSA, prostate volume, and prior biopsy history alongside the imaging findings.

Both scales run from 1 to \rev{5,} and reporting practice may vary between sites and over the study period. Throughout this protocol, \emph{radiologist score} denotes the five-point level of suspicion recorded in the report, whether assigned as a PI-RADS category or as a Likert score, and the scale actually used will be recorded alongside the value for every case. Where the distinction bears on interpretation it is stated explicitly.

\subsection{Data Splitting and Evaluation Paradigm}

\begin{figure}[p]
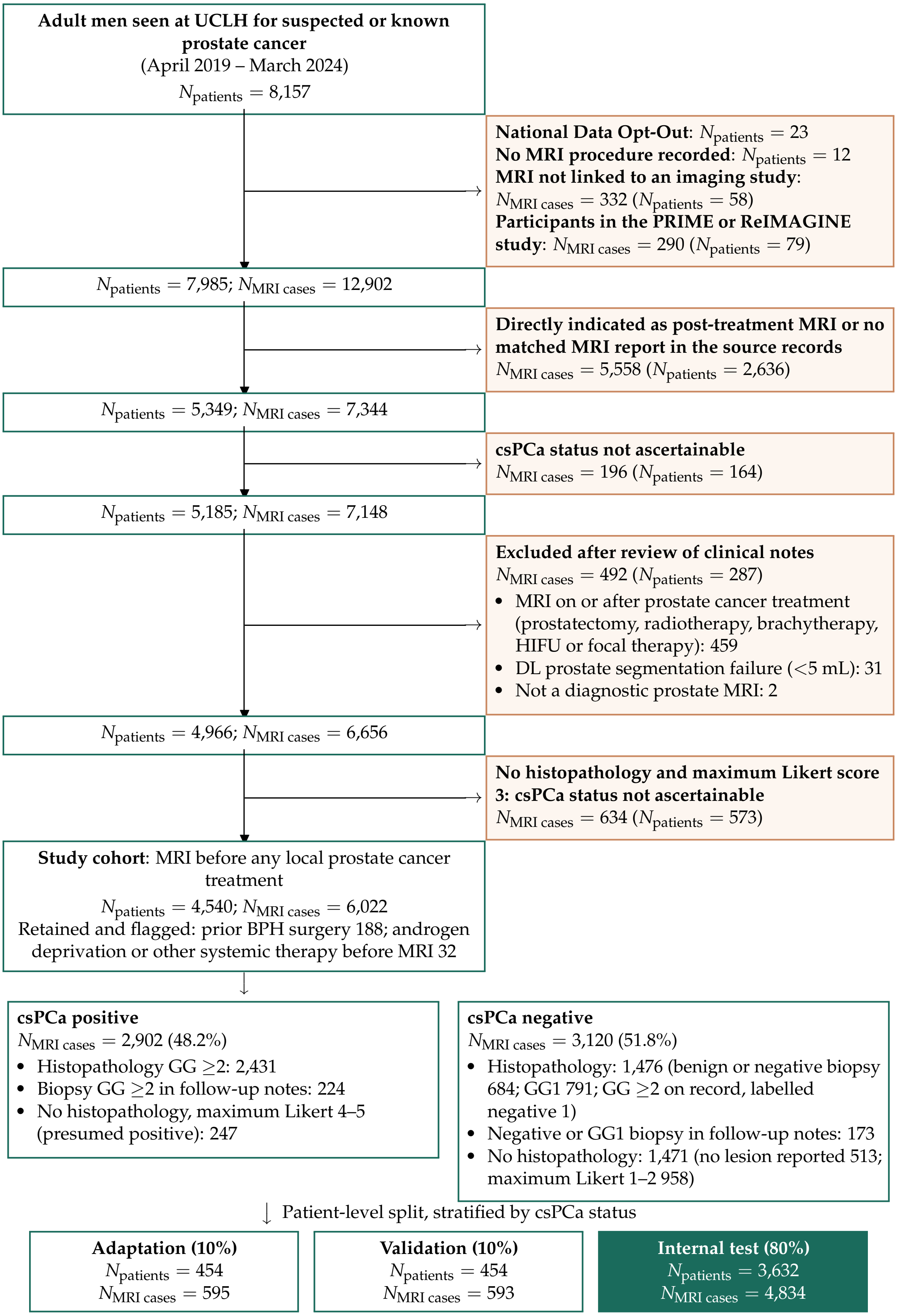

\centering
\footnotesize
\setlength{\fboxsep}{4pt}
\setlength{\fboxrule}{0.8pt}
% \flowbox{content}: a box on the main path. \flowexcl{content}: a side exclusion box; the
% main-path connector is drawn alongside it, sized from the box and the 3pt row gaps so that
% it meets the boxes above and below. The box is set globally because it is measured in one
% table cell and placed in the next.
\newcommand{\flowbox}[1]{\fcolorbox{accentteal}{white}{\parbox{0.50\linewidth}{\centering #1}}}
\newsavebox{\flowexclbox}
\newcommand{\flowexcl}[1]{%
  \global\setbox\flowexclbox=\hbox{\fcolorbox{highlightborder}{warmhighlight}{\parbox[c]{0.42\linewidth}{\raggedright #1}}}%
  \smash{\rule[\dimexpr -\dp\flowexclbox-3pt\relax]{0.6pt}{\dimexpr \ht\flowexclbox+\dp\flowexclbox+6pt\relax}}%
  \makebox[0pt]{\raisebox{\dimexpr -\dp\flowexclbox-2pt\relax}[0pt][0pt]{\scriptsize$\blacktriangledown$}}%
  \makebox[0pt][l]{\rule[\dimexpr 0.25em-0.3pt\relax]{\dimexpr 0.25\linewidth+\fboxsep+\fboxrule-1.1em\relax}{0.6pt}\kern-0.15em$\rightarrow$} &
  \usebox{\flowexclbox}}
\begin{tabular}{@{}c@{}l@{}}

\flowbox{\rev{\textbf{Adult men seen at UCLH for suspected or known prostate cancer}}\\\rev{(April 2019 -- March 2024)}\\[2pt]
\rev{$N_{\text{patients}} = 8{,}157$}} & \\[3pt]

\flowexcl{\rev{\textbf{National Data Opt-Out}: $N_{\text{patients}} = 23$}\\
\rev{\textbf{No MRI procedure recorded}: $N_{\text{patients}} = 12$}\\
\rev{\textbf{MRI not linked to an imaging study}: $N_{\text{MRI cases}} = 332$ ($N_{\text{patients}} = 58$)}\\
\rev{\textbf{Participants in the PRIME or ReIMAGINE study}: $N_{\text{MRI cases}} = 290$ ($N_{\text{patients}} = 79$)}} \\[3pt]

\flowbox{\rev{$N_{\text{patients}} = 7{,}985$; $N_{\text{MRI cases}} = 12{,}902$}} & \\[3pt]

\flowexcl{\rev{\textbf{Directly indicated as post-treatment MRI or no matched MRI report in the source records}}\\\rev{$N_{\text{MRI cases}} = 5{,}558$ ($N_{\text{patients}} = 2{,}636$)}} \\[3pt]

\flowbox{\rev{$N_{\text{patients}} = 5{,}349$; $N_{\text{MRI cases}} = 7{,}344$}} & \\[3pt]

\flowexcl{\rev{\textbf{csPCa status not ascertainable}}\\\rev{$N_{\text{MRI cases}} = 196$ ($N_{\text{patients}} = 164$)}} \\[3pt]

\flowbox{\rev{$N_{\text{patients}} = 5{,}185$; $N_{\text{MRI cases}} = 7{,}148$}} & \\[3pt]

\flowexcl{\rev{\textbf{Excluded after review of clinical notes}}\\\rev{$N_{\text{MRI cases}} = 492$ ($N_{\text{patients}} = 287$)}
\begin{itemize}[leftmargin=1.1em,topsep=2pt,itemsep=0pt,parsep=0pt]
  \item \rev{MRI on or after prostate cancer treatment (prostatectomy, radiotherapy, brachytherapy, HIFU or focal therapy): 459}
  \item \rev{DL prostate segmentation failure ($<$5~mL): 31}
  \item \rev{Not a diagnostic prostate MRI: 2}
\end{itemize}} \\[3pt]

\flowbox{\rev{$N_{\text{patients}} = 4{,}966$; $N_{\text{MRI cases}} = 6{,}656$}} & \\[3pt]

\flowexcl{\rev{\textbf{No histopathology and maximum Likert score 3: csPCa status not ascertainable}}\\\rev{$N_{\text{MRI cases}} = 634$ ($N_{\text{patients}} = 573$)}} \\[3pt]

\flowbox{\rev{\textbf{Study cohort}: MRI before any local prostate cancer treatment}\\[2pt]
\rev{$N_{\text{patients}} = 4{,}540$; $N_{\text{MRI cases}} = 6{,}022$}\\
\rev{Retained and flagged: prior BPH surgery 188; androgen deprivation or other systemic therapy before MRI 32}} & \\[3pt]

$\big\downarrow$ & \\[3pt]

\multicolumn{2}{@{}c@{}}{%
\begin{tabular}{@{}c@{\hspace{1em}}c@{}}
\fcolorbox{accentteal}{white}{\parbox[t]{0.445\linewidth}{\raggedright\textbf{csPCa positive}\\$N_{\text{MRI cases}} = 2{,}902$ (48.2\%)
\begin{itemize}[leftmargin=1.1em,topsep=2pt,itemsep=0pt,parsep=0pt]
  \item Histopathology GG $\geq$2: 2,431
  \item Biopsy GG $\geq$2 in follow-up notes: 224
  \item No histopathology, maximum Likert 4--5 (presumed positive): 247
\end{itemize}}} &
\fcolorbox{accentteal}{white}{\parbox[t]{0.445\linewidth}{\raggedright\textbf{csPCa negative}\\$N_{\text{MRI cases}} = 3{,}120$ (51.8\%)
\begin{itemize}[leftmargin=1.1em,topsep=2pt,itemsep=0pt,parsep=0pt]
  \item Histopathology: 1,476 (benign or negative biopsy 684; GG1 791; GG $\geq$2 on record, labelled negative 1)
  \item Negative or GG1 biopsy in follow-up notes: 173
  \item No histopathology: 1,471 (no lesion reported 513; maximum Likert 1--2 958)
\end{itemize}}}
\end{tabular}} \\[3pt]

\multicolumn{2}{@{}c@{}}{\rev{$\big\downarrow$ \ Patient-level split, stratified by csPCa status}} \\[3pt]

\multicolumn{2}{@{}c@{}}{%
\begin{tabular}{@{}c@{\hspace{0.8em}}c@{\hspace{0.8em}}c@{}}
\fcolorbox{accentteal}{white}{\parbox{0.27\linewidth}{\centering\textbf{Adaptation (10\%)}\\$N_{\text{patients}} = 454$\\$N_{\text{MRI cases}} = 595$}} &
\fcolorbox{accentteal}{white}{\parbox{0.27\linewidth}{\centering\textbf{Validation (10\%)}\\$N_{\text{patients}} = 454$\\$N_{\text{MRI cases}} = 593$}} &
\fcolorbox{accentteal}{accentteal}{\parbox{0.27\linewidth}{\centering\color{white}\textbf{Internal test (80\%)}\\$N_{\text{patients}} = 3{,}632$\\$N_{\text{MRI cases}} = 4{,}834$}}
\end{tabular}} \\
\end{tabular}
\caption{\rev{Flow of patients and MRI cases into the UCLH study cohort, and its division into adaptation, validation, and internal test sets. Exclusion boxes give the patients who had any excluded case, so patient counts in successive boxes need not differ by that number.}}
\rev{\label{fig:flow}}
\end{figure}

The image-only model \rev{is the published PI-CAI-2B system~\citep{SahaSCARLET2026}, with fixed weights; it} was developed \rev{on cohorts external to this study (Section~\ref{sec:image-model}). All development of} the \rev{clinical components uses a UCLH adaptation data set. The UCLH} study \rev{cohort is divided into an adaptation set (10\%), for adapting} the \rev{model to the local setting, a validation set (10\%),} and an \rev{internal} test set \rev{(80\%) held out} for all primary \rev{analyses (Figure~\ref{fig:flow}). External testing uses five further cohorts (Section~\ref{sec:external}).}

% ==========================================================
\section{Study Population}
\label{sec:population}

\subsection{Unit of Analysis}

The unit of analysis is the MRI case: one prostate MRI study together with the clinical data available at the time it was acquired and the csPCa status ascertained for it. A patient may contribute more than one case, since men on active surveillance undergo serial imaging. Eligibility criteria are stated at the level at which they act: criteria concerning the person (age, opt-out status) remove the patient and all of their cases, whereas criteria concerning an individual study (prior treatment, non-diagnostic image quality) remove only that case.

\subsection{Inclusion Criteria}

\begin{enumerate}[label=\arabic*., leftmargin=2em]
  \item Male patients aged $\geq$18 years.
  \item Underwent prostate MRI at UCLH during the study period for initial diagnostic assessment or for active surveillance. Studies performed for staging of already-diagnosed prostate cancer, and studies performed for indications other than prostate cancer detection, are not eligible.
  \item csPCa status ascertainable by the hierarchical labelling procedure of Section~\ref{sec:variables}, which is applied in strict priority order: (a)~histopathology; then (b)~absence of any recorded lesion; then (c)~a maximum radiologist score (PI-RADS or Likert) $<$3; then (d)~an explicitly negative report. A case is labelled by the first applicable step, and a case for which no step applies is excluded.
\end{enumerate}

\noindent Only route~(a) can establish csPCa-positive status; routes~(b)--(d) can establish only a presumed negative. All csPCa-positive cases therefore derive from histopathology, and the negative class is a mixture of biopsy-confirmed and presumed-benign cases. The consequences are addressed in Section~\ref{sec:bias}.

% \noindent Because active surveillance studies are performed in men with an existing diagnosis, the outcome for these cases represents prevalent rather than newly detected disease. The proportion of cases arising from each indication will be reported.

\subsection{Exclusion Criteria}

\begin{enumerate}[label=\arabic*., leftmargin=2em]
  \item MRI studies performed after the patient received \rev{local} prostate cancer treatment (radical prostatectomy, radiotherapy, \rev{brachytherapy, HIFU or focal therapy}). Only studies post-dating the treatment are excluded; earlier studies from the same patient remain eligible. Prior biopsy is not treatment, so studies performed in men on active surveillance who have undergone biopsy alone remain eligible. \rev{MRI after androgen deprivation or other systemic therapy is retained and flagged; because such therapy can reduce gland volume, PSA and tumour conspicuity, a sensitivity analysis excluding these cases is reported.}
  \item Patients who have opted out via the National Data Opt-Out scheme~\citep{NHSEngland2026}.
  \item MRI studies with severe artefact rendering images non-diagnostic, as judged by the reporting radiologist.
\end{enumerate}

% \textbf{Note on verification bias:} Patients with PI-RADS 1-2 scores who do not undergo biopsy are presumed csPCa-negative. This introduces verification bias, which is inherent to the PI-CAI labelling procedure and will be assessed in sensitivity analyses (see Section~\ref{sec:sensitivity}).

\subsection{Data Source}

Structured clinical data are extracted from the UCLH Epic Data Warehouse via the OMOP Extraction System (OMOP ES), a governance-by-default pipeline developed in collaboration between UCL ARC, DHCT, and the UCLH CRDM team. The OMOP ES applies the following safeguards automatically:

\begin{itemize}[leftmargin=2em]
  \item Exclusion of patients registered on the National Data Opt-Out.
  \item Exclusion of sexual health conditions.
  \item Replacement of postcode with Lower Layer Super Output Area (LSOA).
\end{itemize}

\noindent Prostate MRI studies are retrieved from UCLH PACS (Picture Archiving and Communication System) and the Vendor Neutral Archive (VNA). DICOM images are anonymised (removal/pseudonymisation of identifying DICOM tags) and converted to NIfTI format for analysis.

\smallskip
\noindent Figure~\ref{fig:sources} shows the form each source takes and the route by which it reaches the analysis dataset. Date of birth is not retained: the extract carries year of birth only, and age is derived from it and the MRI date. Clinical event dates are retained at full precision because the linkage of pathology to imaging depends on the interval between them (Section~\ref{sec:variables}).

\begin{figure}[p]
\centering
\setlength{\fboxsep}{6pt}\setlength{\fboxrule}{0.7pt}

\begin{center}

\srccard{0.93\linewidth}{Structured extract (OMOP)}{%
person\\[-1pt]\srcrule\\[1pt]
person\_id \ year\_of\_birth \ sex \ ethnicity\_concept \ lsoa\\
P-4821 \ \ \ \ 1958 \ \ \ \ \ \ \ \ \ \ M \ \ \ White British \ \ \ \ E01000882\\[4pt]
measurement\\[-1pt]\srcrule\\[1pt]
person\_id \ measurement\_date \ concept \ \ \ \ \ \ \ \ value \ unit\\
P-4821 \ \ \ \ 2022-03-14 \ \ \ \ \ \ \ PSA \ \ \ \ \ \ \ \ \ \ \ \ 7.4 \ \ \ ng/mL\\
P-4821 \ \ \ \ 2020-11-02 \ \ \ \ \ \ \ PSA \ \ \ \ \ \ \ \ \ \ \ \ 5.1 \ \ \ ng/mL\\
P-4821 \ \ \ \ 2022-02-28 \ \ \ \ \ \ \ Body height \ \ \ \ 178 \ \ \ cm\\
P-4821 \ \ \ \ 2022-02-28 \ \ \ \ \ \ \ Body weight \ \ \ \ 84.2 \ \ kg\\
P-4821 \ \ \ \ 2022-03-01 \ \ \ \ \ \ \ Creatinine \ \ \ \ \ 78 \ \ \ \ umol/L\\
P-4821 \ \ \ \ 2022-03-01 \ \ \ \ \ \ \ Haemoglobin \ \ \ \ 142 \ \ \ g/L\\[4pt]
observation\\[-1pt]\srcrule\\[1pt]
person\_id \ observation\_date \ concept \ \ \ \ \ \ \ \ \ \ \ \ \ value\\
P-4821 \ \ \ \ 2022-02-28 \ \ \ \ \ \ \ Digital rectal exam \ \ Abnormal}

\vspace{2pt}

\srccard{0.93\linewidth}{Histopathology report}{%
SPECIMEN: Prostate, MRI-targeted and systematic needle biopsies\\[2pt]
A. Left apex, target (3 cores, 42 mm total length)\\
\hspace*{1.4em}Acinar adenocarcinoma. Gleason 3+4 = 7 (ISUP grade group 2).\\
\hspace*{1.4em}Pattern 4 component 20\%. Tumour length 9 mm (21\% of cores).\\
\hspace*{1.4em}Perineural invasion present. No extraprostatic extension.\\
B. Left base, systematic (2 cores, 28 mm total length)\\
\hspace*{1.4em}Acinar adenocarcinoma. Gleason 3+3 = 6 (ISUP grade group 1).\\
\hspace*{1.4em}Tumour length 2 mm (7\% of cores).\\
C. Right apex, mid and base, systematic (6 cores, 81 mm)\\
\hspace*{1.4em}Benign prostatic tissue. No evidence of malignancy.\\[2pt]
\srcrule\\[1pt]
SUMMARY: Highest grade Gleason 3+4 = 7, ISUP grade group 2.}

\vspace{2pt}

\begin{tabular}{@{}l@{\hspace{0.7em}}l@{}}
\srccard{0.43\linewidth}{Radiology report}{%
CLINICAL DETAILS: Raised PSA 7.4.\\
Biopsy naive.\\
TECHNIQUE: T2W, DWI (b=0,150,\\
1400), ADC. 3T.\\
FINDINGS: Prostate volume 48 mL.\\
14 mm lesion, left PZ apex.\\
\hl{CONCLUSION: Likert 4. Targeted}\\
\hl{biopsy recommended.}}
&
\srccard{0.43\linewidth}{Clinic letter}{%
Dear Dr \hl{[GP]}, I reviewed\\
\hl{[PATIENT]} in clinic today. He\\
has type 2 diabetes and\\
hypertension, both well\\
controlled. His father was\\
diagnosed with prostate cancer\\
aged 71. He is an ex-smoker\\
(stopped 2009).}
\end{tabular}

\vspace{3pt}$\big\downarrow$\vspace{2pt}

\fcolorbox{highlightborder}{warmhighlight}{\parbox{0.93\linewidth}{\footnotesize
\begin{tabular}{@{}L{3.1cm}L{3.4cm}L{7.0cm}@{}}
\textbf{Source} & \textbf{Route} & \textbf{Variables obtained} \\
\midrule
Structured extract & Used directly & PSA, PSA change, height, weight, BMI, BMI change, DRE finding, ethnicity, laboratory results, medication \\
\addlinespace[2pt]
Radiology report & LLM extraction & Radiologist score, lesion counts, lesion location, PSA, prostate volume \\
\addlinespace[2pt]
Histopathology report & LLM extraction & Gleason score, ISUP grade group \\
\addlinespace[2pt]
Clinic letters & LLM extraction & Comorbidities, smoking, alcohol, family history, history of other cancers, prior BPH intervention \\
\addlinespace[2pt]
DICOM (PACS / VNA) & Preprocessing, segmentation, intensity classification & Image channels, prostate volume, PPAT measurements, retropubic gap \\
\addlinespace[2pt]
Combinations of the above & Derived & Age (MRI date minus year of birth), PSA density, PSA change (MRI report) \\
\end{tabular}}}
\end{center}

\caption{Data sources and the route by which each reaches the analysis dataset. The excerpts are illustrative synthetic examples, not real patient records. Shading marks the report section to which the keyword rules of Table~\ref{tab:supp_keywords} apply, and the placeholders substituted during pseudonymisation.}
\label{fig:sources}
\end{figure}

% ==========================================================
\section{Variables and Outcomes}
\label{sec:variables}

\subsection{Primary Outcome}

Presence of clinically significant prostate cancer (csPCa), defined following the PI-CAI challenge labelling procedure~\citep{Saha2024} as a case-level binary label (0 = no csPCa; 1 = csPCa) derived using the following hierarchical procedure:

\begin{enumerate}[label=\arabic*., leftmargin=2em]
  \item \textbf{Histopathology (highest priority):} If a Gleason score from prostate biopsy or radical prostatectomy is available, the case is labelled csPCa-positive when Gleason score $\geq$7 (i.e., ISUP grade group $\geq$2; Gleason $\geq$3+4) and csPCa-negative when Gleason score $<$7.
  \item \textbf{Lesion count:} If no histopathology is available and the radiology report records no lesion, the case is labelled csPCa-negative.
  \item \textbf{Radiologist score:} If neither histopathology nor lesion count is available and the maximum radiologist score (PI-RADS or Likert) is $<$3, the case is labelled csPCa-negative. This follows the PI-CAI assumption that cases scored 1--2 without biopsy are presumed benign, applied here to whichever scale the report used.
  \item \textbf{Radiology report keywords:} If the above conditions are unmet, cases with explicitly negative radiology reports (e.g., ``no suspicious lesion'', ``screen negative'') are labelled csPCa-negative. The complete rule set, including the exclusion patterns that prevent a label being assigned where the report expresses diagnostic uncertainty, is given in Table~\ref{tab:supp_keywords}.
\end{enumerate}

\paragraph{Linkage of pathology to imaging.}
A histopathology result is linked to an MRI case only where the specimen was obtained within 3~months after the MRI and before any intervening prostate cancer treatment. Where a patient has several MRI studies and several specimens, each specimen is linked to the most recent preceding eligible MRI, and each MRI case takes the earliest specimen linked to it. An MRI case with no specimen inside the window is labelled by the remaining steps of the hierarchy rather than by a later specimen, so that disease detected long after the index study is not attributed to it. The sensitivity of the results to this interval is examined in Section~\ref{sec:sensitivity}.

\paragraph{Alignment with PI-CAI.}
This procedure aligns with the labelling rules of the PI-CAI challenge~\citep{Saha2024}: the priority given to histopathology, the ISUP grade group $\geq$2 threshold, and the presumption that unbiopsied cases scored 1--2 are benign are all adopted unchanged.\footnote{Three adaptations are required by the UCLH data sources. First, PI-CAI was defined for cohorts reported using PI-RADS, whereas the score here may be a PI-RADS category or a Likert score depending on local reporting practice (Section~\ref{sec:design}). Second, a histopathology result is linked to an MRI case only within a defined interval, whereas PI-CAI applied no explicit window; this makes labelling stricter, and shifts some cases from histopathological to presumed-negative labelling. Third, a keyword step is added for reports that state absence of disease without recording a score; this makes labelling more permissive, and recovers cases that would otherwise be unclassifiable and excluded.}

\subsection{Imaging Predictors}
\label{sec:imaging}

Prostate MRI studies at UCLH are acquired on 1.5T and 3T scanners as part of routine clinical care. The bi-parametric MRI protocol comprises three input channels.

\begin{table}[htbp]
\centering
\caption{Bi-parametric MRI sequences used as imaging predictors.}
\label{tab:imaging}
\begin{tabular}{@{}L{3.8cm}L{4.0cm}L{4.5cm}@{}}
\toprule
\textbf{Sequence} & \textbf{Purpose} & \textbf{Key Parameters} \\
\midrule
T2-weighted \rev{imaging} (T2W) & Anatomical and zonal assessment and lesion morphology & High-resolution T2W imaging \\
High \rev{$b$-value diffusion-weighted imaging (HBV)} & Detection of restricted diffusion and lesion conspicuity & High \rev{$b$-value} $\geq$1400~s/mm$^2$ \\
\rev{Apparent diffusion coefficient (ADC)} map & Quantitative assessment of diffusion restriction & Calculated from \rev{diffusion-weighted imaging (DWI)} acquired at multiple \rev{$b$-values} \\
\bottomrule
\end{tabular}
\end{table}

\paragraph{Image preprocessing.}
\begin{enumerate}[label=\arabic*., leftmargin=2em]
  \item \textbf{Format conversion:} DICOM series are converted to NIfTI using \texttt{dcm2niix}. Manufacturer rescaling is applied at conversion so that ADC maps are expressed in $\times 10^{-6}$~mm$^2$/s irrespective of vendor; series whose scaling cannot be resolved are excluded. The high $b$-value series is identified from the $b$-value recorded in the standard or vendor-private tags, and whether it was acquired or computed is recorded for each case.

  \item \textbf{Anonymisation:} Identifying tags are removed or pseudonymised following the Basic Application Level Confidentiality Profile of DICOM PS3.15 Annex~E. Series liable to carry burned-in annotation, such as secondary captures, are excluded.

  \item \textbf{Spatial standardisation:} T2W, high $b$-value DWI, and ADC are resampled into a common reference frame at a fixed voxel spacing of $0.5 \times 0.5 \times 3.0$~mm, to a uniform matrix of $256 \times 256 \times 20$ about the image centre. \rev{This reference frame is used for the image-derived measurements (Section~\ref{sec:image-derived}); the input to the image model is prepared separately in step~4.}

  \item \rev{\textbf{Model-specific preprocessing:} The input to PI-CAI-2B is prepared with the preprocessing released with the model~\citep{SahaSCARLET2026}: the prostate is segmented on T2W; ADC} and high $b$-value DWI are \rev{registered to T2W; the volumes are cropped to the prostate bounding box with a 10~mm margin, and resampling and intensity normalisation follow the} nnU-Net \rev{plans of} the \rev{model~\citep{Isensee2021}. The preprocessing} is identical to that used to train \rev{PI-CAI-2B,} necessary for using \rev{its fixed weights} (Section~\ref{sec: image encoder}).

  \item \textbf{Quality control:} Converted volumes are checked for correct orientation, plausible geometry, and non-empty content. In addition, the median ADC within the segmented prostate is computed per scanner and compared against a physiologically plausible range of approximately 1{,}000--1{,}800 $\times 10^{-6}$~mm$^2$/s; a systematic departure indicates unresolved vendor scaling rather than a biological difference. Failures are listed and excluded.
\end{enumerate}

\subsection{Clinical Predictor Variables}

\begin{longtable}{@{}L{3.2cm}L{2.7cm}L{2.2cm}L{2.0cm}L{2.6cm}@{}}
\caption{Clinical predictor variables, data sources, types, and preprocessing.}
\label{tab:clinical_vars} \\
\toprule
\textbf{Variable} & \textbf{Source} & \textbf{Type} & \textbf{Transform} & \textbf{Normalisation} \\
\midrule
\endfirsthead
\toprule
\textbf{Variable} & \textbf{Source} & \textbf{Type} & \textbf{Transform} & \textbf{Normalisation} \\
\midrule
\endhead
\bottomrule
\endfoot
\addlinespace[4pt]
\multicolumn{5}{@{}l@{}}{\rev{\textbf{I. Patient context}}} \\*
\addlinespace[2pt]
Age & Structured \rev{EHR} & Continuous & None & Z-score \\
\rev{Ethnicity} & \rev{Structured EHR} & \rev{Categorical (6)} & \rev{One-hot encoded} & \rev{None} \\
\rev{Height (cm)} & \rev{Structured EHR} & \rev{Continuous} & \rev{None} & \rev{Z-score} \\
\rev{Weight (kg)} & \rev{Structured EHR} & \rev{Continuous} & \rev{None} & \rev{Z-score} \\
\rev{BMI (kg/m$^2$)} & \rev{Structured EHR} & \rev{Continuous} & \rev{$\log(x+1)$} & \rev{Z-score} \\
\rev{BMI change} & \rev{Structured EHR} & \rev{Continuous} & \rev{None} & \rev{Z-score} \\
\rev{Smoking status} & \rev{LLM} & \rev{Categorical (3)} & \rev{One-hot encoded} & \rev{None} \\
\rev{Alcohol status} & \rev{LLM} & \rev{Categorical (7)} & \rev{One-hot encoded} & \rev{None} \\
\rev{Family hx any cancer} & \rev{LLM} & \rev{Binary} & \rev{Binary encoded (0/1)} & \rev{None} \\
\rev{Family hx prostate ca.} & \rev{LLM} & \rev{Binary} & \rev{Binary encoded (0/1)} & \rev{None} \\
\rev{Family hx breast ca.} & \rev{LLM} & \rev{Binary} & \rev{Binary encoded (0/1)} & \rev{None} \\
\rev{Family hx ovarian ca.} & \rev{LLM} & \rev{Binary} & \rev{Binary encoded (0/1)} & \rev{None} \\
\rev{Hx other cancers} & \rev{LLM} & \rev{Binary} & \rev{Binary encoded (0/1)} & \rev{None} \\
\rev{Diabetes} & \rev{LLM} & \rev{Binary} & \rev{Binary encoded (0/1)} & \rev{None} \\
\rev{Hypertension} & \rev{LLM} & \rev{Binary} & \rev{Binary encoded (0/1)} & \rev{None} \\
\rev{Cardiovascular disease} & \rev{LLM} & \rev{Binary} & \rev{Binary encoded (0/1)} & \rev{None} \\
\rev{COPD} & \rev{LLM} & \rev{Binary} & \rev{Binary encoded (0/1)} & \rev{None} \\
\rev{Asthma} & \rev{LLM} & \rev{Binary} & \rev{Binary encoded (0/1)} & \rev{None} \\
\rev{Chronic kidney disease} & \rev{LLM} & \rev{Binary} & \rev{Binary encoded (0/1)} & \rev{None} \\
\rev{DRE finding} & \rev{Structured EHR} & \rev{Binary} & \rev{Binary encoded (0/1)} & \rev{None} \\
\rev{PSA, laboratory} (ng/mL) & Structured \rev{EHR} & Continuous & $\log(x+1)$ & Z-score \\
PSA \rev{change, laboratory} & Structured \rev{EHR} & Continuous & None & \rev{Z-score} \\
\rev{PSA, MRI report (ng/mL)} & \rev{LLM} & \rev{Continuous} & \rev{$\log(x+1)$} & \rev{Z-score} \\
\rev{PSA change, MRI report} & \rev{Derived} & \rev{Continuous} & \rev{None} & \rev{Z-score} \\
\rev{Blood} & \rev{Structured EHR} & \rev{Binary} & \rev{Binary encoded (0/1)} & \rev{None} \\
\rev{Urine} & \rev{Structured EHR} & \rev{Binary} & \rev{Binary encoded (0/1)} & \rev{None} \\
\rev{5-ARI use at MRI} & \rev{LLM} & \rev{Binary} & \rev{Binary encoded (0/1)} & \rev{None} \\
\rev{Aspirin / NSAID use at MRI} & \rev{LLM} & \rev{Binary} & \rev{Binary encoded (0/1)} & \rev{None} \\
\rev{Prior BPH intervention} & \rev{LLM} & \rev{Binary} & \rev{Binary encoded (0/1)} & \rev{None} \\
\addlinespace[4pt]
\multicolumn{5}{@{}l@{}}{\rev{\textbf{II. Radiologist-derived context}}} \\*
\addlinespace[2pt]
\rev{Radiologist score (1--5)} & \rev{LLM} & \rev{Ordinal} & \rev{None} & \rev{None (raw)} \\
\rev{MRI lesion count} & \rev{LLM} & \rev{Count} & \rev{$\log(x+1)$} & \rev{Z-score} \\
\rev{Lesion count with radiologist score $\geq$3} & \rev{LLM} & \rev{Count} & \rev{$\log(x+1)$} & \rev{Z-score} \\
\rev{Overall lesion laterality} & \rev{LLM} & \rev{Categorical} & \rev{One-hot encoded} & \rev{None} \\
\rev{Index lesion side} & \rev{LLM} & \rev{Categorical} & \rev{One-hot encoded} & \rev{None} \\
\rev{Index lesion zone} & \rev{LLM} & \rev{Multi-label categorical} & \rev{Multi-hot encoded} & \rev{None} \\
\rev{Index lesion level} & \rev{LLM} & \rev{Multi-label categorical} & \rev{Multi-hot encoded} & \rev{None} \\
\addlinespace[4pt]
\multicolumn{5}{@{}l@{}}{\rev{\textbf{III. Quantitative imaging context}}} \\*
\addlinespace[2pt]
\rev{Prostate volume (mL)} & \rev{DL segmentation} & \rev{Continuous} & \rev{$\log(x+1)$} & Z-score \\
PSA density (ng/mL/mL) & Derived & Continuous & $\log(x+1)$ & Z-score \\
\rev{PPAT} volume (mL) & DL segmentation & Continuous & $\log(x+1)$ & Z-score \\
\rev{PPAT / prostate volume ratio} & \rev{Derived} & Continuous & $\log(x+1)$ & Z-score \\
\rev{PPAT thickness (mm)} & \rev{Derived} & Continuous & None & Z-score \\
\rev{PPAT fat fraction} & \rev{Derived} & \rev{Continuous (0--1)} & \rev{None} & \rev{Z-score} \\
\rev{Retropubic gap (mm)} & \rev{Derived} & Continuous & None & Z-score \\
\end{longtable}

\noindent\textbf{Source key:} \rev{\emph{Structured EHR}} = extracted directly from structured fields in the Epic EHR (OMOP), including structured Parquet exports; \emph{DL segmentation} = prostate volume computed from automated deep learning segmentation of the MRI; \emph{Derived} = calculated from other variables (PSA density = PSA / prostate volume); \emph{LLM} = extracted from unstructured text (radiology reports, histopathology reports, and clinical notes) using the large language model pipeline described in Section~\ref{sec:llm-extraction}.

\subsection{Missing Data Handling}
\label{sec: missing data}

\rev{In the image-context models, missing values are not imputed. Each} clinical \rev{variable $j$ carries an indicator $m_j$, equal to 1 when the variable is recorded and 0 when it is missing, and its contribution to the model is multiplied by $m_j$ (Section~\ref{sec: clinical encoder}). A missing variable therefore contributes nothing, and} a \rev{case with no recorded variable receives exactly} the \rev{image-only prediction. Image-derived measurements that fail quality control, or that cannot be measured for a case (Section~\ref{sec:image-derived}),} are \rev{treated as missing in} the \rev{same way.}

\subsection{LLM-Based Variable Extraction}
\label{sec:llm-extraction}

Several clinical variables listed in Table~\ref{tab:clinical_vars} are not available as structured data fields and must be extracted from unstructured free-text clinical documents. A large language model (LLM)-based natural language processing (NLP) pipeline will be developed for this purpose. The pipeline is designed to maximise extraction reliability and auditability through the following stages.

\paragraph{1.\ Source document identification and preparation.}
Three categories of unstructured text will be used: (i)~radiology reports, from which radiologist \rev{scores, lesion counts, lesion location (overall laterality,} and \rev{the side, zone, and level of the index lesion), and the PSA value and prostate volume stated in the report} are extracted; (ii)~histopathology reports, from which Gleason scores and ISUP grade groups are extracted; and (iii)~clinical notes (consultation letters, clinic letters, referral letters, and pre-operative assessments), from which comorbidities, lifestyle factors, family \rev{history (including breast} and \rev{ovarian cancer),} history of other \rev{cancers, medication use at the time of MRI (5-ARI, aspirin or NSAID), and prior BPH intervention} are extracted. Variables recorded in structured fields, including anthropometric measurements and digital rectal examination findings, are taken from the structured extract (Table~\ref{tab:clinical_vars}); prostate volume \rev{and the PPAT measurements are} computed from \rev{the MRI (Section~\ref{sec:image-derived}).} All documents are exported from the OMOP-mapped EHR, split into per-patient corpora, and pseudonymised (removal of patient names, NHS numbers, dates of birth, and free-text addresses) prior to LLM processing.

\paragraph{2.\ Prompt design and extraction schema.}
Each patient's documents are submitted to the LLM with a structured JSON schema prompt that (a)~enumerates every target variable with its permissible data type and value set, (b)~mandates explicit \texttt{null} returns for any variable not mentioned in the source text, preventing the model from inferring or fabricating values, and (c)~includes task-specific instructions (e.g., ``distinguish between cancer-related and non-cancer family history'', ``do not infer a diagnosis from medication alone''). Prompts are developed iteratively on a held-out development set of 50 patients and frozen before production extraction begins. The system prompt and the per-variable extraction schema are reproduced in full in Section~\ref{sec:supp-schema} and Table~\ref{tab:supp_schema}.

\paragraph{3.\ Model selection and execution.}
The extraction will use a commercial LLM accessed via API (model selection to be finalised based on benchmark performance on the development set; candidates including Claude Sonnet 4.5, Gemini 2.5 Pro, and GPT-5.2). The selected model will process each patient's notes in a zero-shot setting with the frozen prompt. To ensure reproducibility, the model version, API parameters (temperature = 0, maximum output tokens), and prompt text will be recorded and version-controlled. An automated execution pipeline will handle rate-limiting, retry logic (exponential back-off with a maximum of three retries per patient), and graceful degradation for notes exceeding the model's context window (truncation with a trailing marker).

\paragraph{4.\ Output parsing and post-processing.}
Raw LLM outputs will be parsed as JSON with the following safeguards: (i)~responses wrapped in markdown code fences are automatically unwrapped; (ii)~malformed JSON triggers one retry with an explicit ``return valid JSON only'' instruction; (iii)~parsed outputs are validated against the extraction schema (type checking, range checking for numerical fields, controlled vocabulary matching for categorical fields). After parsing, rule-based post-processing standardises free-text entries - for example, cancer type descriptions are mapped to a controlled vocabulary using regular-expression rules, and benign conditions erroneously included in cancer fields are filtered out. Internal consistency checks are applied (e.g., if family history of prostate cancer is flagged as true, ``prostate cancer'' must appear in the cancer types list). \rev{Smoking status and} alcohol \rev{status} are \rev{extracted as integer codes (Table~\ref{tab:supp_schema}) and one-hot encoded for modelling (Table~\ref{tab:clinical_vars}). One-hot encoding discards the order of the seven alcohol bands but allows a non-monotonic association with csPCa, since the dose--response relation between alcohol and prostate cancer may be non-linear~\citep{Hong2020}.} Figure~\ref{fig:extraction} works through the sequence for one document, including a value that is discarded because its quotation cannot be located.

\begin{figure}[p]
\centering
\setlength{\fboxsep}{6pt}\setlength{\fboxrule}{0.7pt}

\begin{center}

\srccard{0.93\linewidth}{1. Source document, after pseudonymisation}{%
Dear Dr \hl{[GP]},\\[2pt]
I reviewed \hl{[PATIENT]} in clinic today following his recent MRI. He has\\
\hl{type 2 diabetes} and hypertension, both well controlled on metformin and\\
ramipril. \hl{There is no history of cardiovascular disease.} His father was\\
diagnosed with prostate cancer aged 71. He \hl{stopped smoking in 2009}.}

\dn

\srccard{0.93\linewidth}{2. Extraction schema submitted with the document (extract)}{%
"diabetes": \{\\
\hspace*{1.4em}"type": ["boolean", "null"],\\
\hspace*{1.4em}"rule": "true where recorded as a current or past diagnosis;\\
\hspace*{4.6em}false only where explicitly denied; null where not\\
\hspace*{4.6em}mentioned. Do not infer from medication alone."\\
\},\\
"diabetes\_evidence": \{ "type": ["string", "null"] \}}

\dn

\srccard{0.93\linewidth}{3. Returned JSON}{%
\{\\
\hspace*{1.4em}"diabetes": true,\\
\hspace*{1.4em}"diabetes\_evidence": "type 2 diabetes",\\
\hspace*{1.4em}"cardiovascular\_disease": false,\\
\hspace*{1.4em}"cardiovascular\_disease\_evidence": "no history of cardiovascular disease",\\
\hspace*{1.4em}"chronic\_kidney\_disease": null,\\
\hspace*{1.4em}"chronic\_kidney\_disease\_evidence": null,\\
\hspace*{1.4em}"smoking\_status": 1,\\
\hspace*{1.4em}"smoking\_status\_evidence": "stopped smoking in 2009",\\
\hspace*{1.4em}\bad{"copd": true,}\\
\hspace*{1.4em}\bad{"copd\_evidence": "long-standing COPD"}\\
\}}

\dn

\fcolorbox{highlightborder}{warmhighlight}{\parbox{0.93\linewidth}{\footnotesize
\textbf{4. Validation against the source, then the value entering the dataset}\\[4pt]
\begin{tabular}{@{}L{4.0cm}C{3.4cm}L{6.0cm}@{}}
\textbf{Field} & \textbf{Quotation located} & \textbf{Value retained} \\
\midrule
diabetes & yes & \texttt{true} \\
cardiovascular\_disease & yes & \texttt{false} (explicitly denied) \\
chronic\_kidney\_disease & not applicable & \texttt{null} (not mentioned) \\
smoking\_status & yes & \texttt{1} (former smoker) \\
\addlinespace[2pt]
copd & \textbf{no} & \texttt{null}; the returned value is discarded and the case flagged \\
\end{tabular}}}

\end{center}

\caption{How a value is extracted from free text and what prevents one being invented. The document is an illustrative synthetic example, not a real patient record. Each non-null value must be returned with a supporting quotation, which is then located in the source document after normalisation of case and whitespace; a value whose quotation cannot be found is discarded rather than accepted, as shown for \texttt{copd}. The three states \texttt{true}, \texttt{false} and \texttt{null} are distinct: absence of mention is \texttt{null}, never \texttt{false}.}
\label{fig:extraction}
\end{figure}

\paragraph{5.\ Duplicate extraction and stability assessment.}
A random 10\% sample of patients (minimum 300) will undergo duplicate extraction: the frozen prompt will be run a second time with the primary model, and once with a second model from the candidate set. Per-variable agreement between runs will be quantified using Cohen's $\kappa$ for categorical variables and the intraclass correlation coefficient for continuous variables. Two forms of instability are distinguished. Run-to-run disagreement under the same model indicates residual non-determinism despite temperature~0; variables falling below $\kappa=0.90$ will be examined for prompt ambiguity and the prompt revised before it is frozen. Between-model disagreement indicates model-specific interpretation of the prompt; it is reported descriptively and used to bound the model-dependence of downstream results. This assessment is performed on the development set before the prompt is frozen, so it does not consume the validation sample described below.

\paragraph{6.\ Validation against manual chart review.}
A random sample of at least 200 patients (stratified by csPCa status) will undergo independent manual chart review by two clinically trained annotators. Inter-annotator agreement will be quantified using Cohen's $\kappa$. LLM extractions will then be compared against the consensus annotation to compute per-variable precision, recall, and F$_1$ score. Variables with F$_1 < 0.80$ will be flagged; depending on the nature of the errors, these variables will be either excluded from analysis, subjected to additional manual correction, or retained with a sensitivity analysis assessing their impact on model performance.

\paragraph{7.\ Adjudicated re-review of discordant variables.}
Cases in which the LLM output and the two annotators do not all agree, together with all variables failing the F$_1$ threshold, will undergo a second review in a smaller adjudication set of 50 patients drawn from the validation sample. Review is performed by a third clinically trained reviewer, blinded to both the LLM output and the initial annotations, who assigns an independent value from the source documents; this adjudicated value becomes the reference standard for those cases. Each discrepancy is classified as source ambiguity, annotator error, or extraction error, and counts by category are reported in Table~\ref{tab:supp_agreement}. The classification matters because only extraction error is attributable to the pipeline and correctable by prompt revision, whereas source ambiguity sets a ceiling on the accuracy any extraction method could reach for that variable.

\paragraph{8.\ Auditability and traceability.}
Every extracted value will be linked to its source document identifier, enabling downstream auditing. The full extraction pipeline - prompt templates, model version, post-processing rules, and validation results - will be documented and released as supplementary material to support reproducibility.

\paragraph{9.\ Handling of extraction failures.}
Patients for whom extraction fails after all retries (e.g., persistent API errors, unparseable responses) will be coded as missing across all LLM-derived variables. These missing values are handled \rev{as} described in Section~\rev{\ref{sec: missing data}}.

\subsection{\rev{Image-Derived Variable Measurement}}
\rev{\label{sec:image-derived}}

\rev{Prostate volume and the periprostatic adipose tissue (PPAT) measurements in Table~\ref{tab:clinical_vars} are computed automatically from the T2W series. Like the variables extracted from free text (Section~\ref{sec:llm-extraction}), they enter the image-context model as scalar clinical predictors. They differ from the imaging predictors of Section~\ref{sec:imaging}, which are image channels passed to the image encoder. All measurements are computed on the resampled T2W volume (Section~\ref{sec:imaging}, step~3) before cropping, which covers a $128 \times 128 \times 60$~mm field of view centred on the prostate.}

\paragraph{\rev{Anatomical segmentation.}}
\rev{The prostate and surrounding structures (bladder, rectum, bowel, pelvic bones, iliac vessels, and pelvic muscles) are segmented with the pretrained TotalSegmentator MRI models~\citep{Wasserthal2023,DAntonoli2025}, without retraining or fine-tuning on UCLH data.}

\paragraph{\rev{Adipose tissue classification.}}
\rev{No pretrained model for segmenting PPAT on T2W images was available, so adipose tissue, the high-signal class on non-fat-suppressed T2W images, is classified by signal intensity: after N4 bias-field correction~\citep{Tustison2010}, a threshold is set for each case by Otsu's method~\citep{Otsu1979}, and segmented non-adipose structures are removed. Because the threshold is set on an arbitrary intensity scale, it does not transfer to other acquisition protocols.}
% Implementation details moved out of the protocol (for the paper):
% The T2W series is not fat-suppressed, so fat is the high-signal tissue class. The Otsu threshold is computed over the field of view, excluding the bladder, whose urine signal exceeds that of fat.
% To guard against a degenerate split, the threshold is constrained to lie between 0.45 and 0.90 times the 80th percentile of signal in the segmented pelvic bone, which consists predominantly of fatty marrow.
% A voxel is classified as adipose if its signal reaches the threshold and it does not belong to any segmented non-adipose structure.
% The threshold is reproducible within this cohort.

\paragraph{\rev{Periprostatic adipose tissue.}}
\rev{PPAT is included as an additional, exploratory predictor. It has been hypothesised that immune-active cells within PPAT regulate disease development and progression~\citep{Gucalp2017,Finley2009}, and periprostatic fat thickness on MRI has been associated with high-grade prostate cancer~\citep{Cao2017}. PPAT is defined as adipose tissue within 15~mm of the prostate surface over the axial extent of the gland; bands of 10~mm and 20~mm are used in sensitivity analysis (Section~\ref{sec:sensitivity}).}
% Implementation details moved out of the protocol (for the paper):
% The 15 mm band is measured by a three-dimensional distance transform that accounts for the anisotropic voxel spacing, and is restricted to the axial slices spanned by the gland.

\paragraph{\rev{Measurements.}}
\rev{Six measurements are derived from these automated segmentations (Table~\ref{tab:clinical_vars}, group~III); PSA density is then calculated from prostate volume.}
\begin{itemize}[leftmargin=2em]
  \item \rev{\emph{Prostate volume}: the volume of the prostate mask, except in truncated cases (see below).}
  \item \rev{\emph{PPAT volume}: the volume of PPAT at 15~mm.}
  \item \rev{\emph{PPAT / prostate volume ratio}: PPAT volume divided by prostate volume.}
  \item \rev{\emph{PPAT fat fraction}: the share of the 15~mm periprostatic band, after removal of other segmented structures, that is classified as adipose. It is a proportion of voxels classified by signal intensity, not a physical fat fraction measured with a fat-specific sequence.}
  \item \rev{\emph{Retropubic gap}: on the mid-gland axial slice, within a strip 10~mm either side of the gland midline, the anteroposterior distance from the anterior surface of the prostate to the posterior surface of the pubic bone. Where no pubic bone is segmented anterior to the gland on that slice, up to two adjacent slices on each side are used instead.}
  \item \rev{\emph{PPAT thickness}: the adipose depth within the retropubic gap, equal to the gap multiplied by the proportion of the strip classified as adipose.}
\end{itemize}

\paragraph{\rev{Quality control.}}
\rev{The image-derived measurements of a case are set to missing if the prostate mask is empty, is fragmented into disconnected components, or has a volume outside 10--250~mL; if PPAT volume is below 0.5~mL; or if the PPAT fat fraction lies outside 0.02--0.95. The retropubic gap and PPAT thickness are also set to missing where no pubic bone is segmented anterior to the gland. Overlay images of the masks at apex, mid-gland, and base are generated for every case for visual review. The case itself remains in the cohort, and the number of cases failing each check is reported.}

\paragraph{\rev{Gland truncation.}}
\rev{The 60~mm craniocaudal extent of the field of view does not cover the whole gland in larger prostates, so the segmented volume underestimates the true volume. Cases in which the prostate mask reaches the superior or inferior boundary of the field of view are flagged, and their number is reported. For these cases, prostate volume, and therefore PSA density, is taken from the volume stated in the radiology report; where the report states no volume, the segmented volume is retained. PPAT volume and the measurements derived from it remain underestimated in flagged cases, and their effect is examined in a sensitivity analysis (Section~\ref{sec:sensitivity}).}

\paragraph{\rev{Validation.}}
\rev{Prostate volume from segmentation is compared with the volume stated in the radiology report in all cases where a volume is stated, using the intraclass correlation coefficient and Bland--Altman analysis. No reference PPAT measurement exists in routine reports, and manual annotation is labour-intensive, so PPAT segmentation is compared with manual annotation by a trained reader in a random sample of 50 cases, stratified by scanner and gland size, reporting the Dice coefficient and relative volume error.}

\subsection{Data Linkage and Storage}

All data sources are linked at the patient level using pseudonymised identifiers generated during the OMOP extraction process. An auditable list of all data items will be maintained, with statistical disclosure control applied to prevent re-identification through unique attribute constellations. Data are stored in the UCL Data Safe Haven or UCLH Azure landing zones with role-based access control (RBAC). GPU computing resources are provisioned within these secure environments for model training.

% ==========================================================
\section{Sample Size}
\label{sec:sample}

The study is expected to include approximately 7,171 eligible MRI cases from 5,208 patients with linked clinical data, of whom a subset will have corresponding preprocessed MRI images available for multimodal analysis.

\begin{table}[htbp]
\centering
\caption{Summary of cohort characteristics. Counts and the age distribution are given per MRI case unless stated otherwise.}
\label{tab:cohort}
\begin{tabular}{@{}lr@{}}
\toprule
\textbf{Characteristic} & \textbf{Value} \\
\midrule
Total eligible MRI cases & 7,171 \\
Total patients & 5,208 \\
csPCa positive (ISUP $\geq$2) & 2,448 (34.1\%) \\
csPCa negative & 4,723 (65.9\%) \\
Median age at MRI (IQR) & 66 (60-73) years \\
\bottomrule
\end{tabular}
\end{table}

\subsection{Sample Size Justification}

Formal sample size calculations for prediction models follow the method of Riley et al.~\citep{Riley2020}. With an anticipated outcome prevalence of 34\%, $\geq$20 candidate predictor parameters (imaging features + clinical variables), a target Cox-Snell $R^2$ of 0.15, and a maximum allowable shrinkage factor of 0.9, the minimum sample size required is approximately 1,200 patients. Our available cohort of $>$7,000 patients substantially exceeds this minimum, providing adequate statistical power for both primary analysis and subgroup evaluations.

% ==========================================================
\section{Statistical Analysis Plan}
\label{sec:analysis}

\subsection{Descriptive Statistics}

Baseline characteristics will be summarised by csPCa status, using means (SD) or medians (IQR) for continuous variables and frequencies (\%) for categorical variables. Missingness will be reported per variable and separately for the \rev{adaptation, validation,} and test \rev{sets,} since missingness rates range from 8\% to 78\% across the clinical predictors (Table~\ref{tab:bias}).

\subsection{\rev{Image-only Model and Data Splitting}}
\rev{\label{sec:image-model}}

The image-only model \rev{is PI-CAI-2B, an AI system for the detection of csPCa (ISUP grade group $\geq$2) on bpMRI,} developed \rev{and validated in} the \rev{SCARLET-1 study~\citep{SahaSCARLETProtocol2025,SahaSCARLET2026}. It is a further development of the top-performing algorithm of the PI-CAI study~\citep{Saha2024}: a three-dimensional nnU-Net (v1) model~\citep{Isensee2021} with a MedNeXt-3XL encoder--decoder of approximately $1.0 \times 10^{9}$ parameters, trained as five cross-validation folds whose outputs are averaged at inference. It takes T2W, ADC, and high $b$-value DWI as input and returns a voxel-level csPCa likelihood map.}

\rev{The released system can combine its image score with clinical information. This} study \rev{uses fold~0 of the released weights in the image-only configuration, in which no clinical variable enters the model. The case-level score is derived} from the \rev{likelihood map alone: voxel likelihoods below 0.1 are set to zero, each connected lesion candidate is assigned its mean likelihood,} and \rev{the case score is} the \rev{highest} lesion \rev{likelihood. Because} the image-only model \rev{uses no clinical information, it is the reference against which the contribution of clinical variables is measured.}

\paragraph{\rev{Development and test cohorts of PI-CAI-2B.}} \rev{As stated in the SCARLET-1 study protocol~\citep{SahaSCARLETProtocol2025}, PI-CAI-2B was trained on 18,371 MRI examinations (17,178 patients; 26 cities in 14 countries) drawn from two cohorts:}

\begin{itemize}[leftmargin=2em]
  \item \rev{\textbf{Primary training and internal testing cohort:} 10,207 examinations (9,129 patients) acquired from 2012 to 2021 at four tertiary centres in the Netherlands} and \rev{Norway, originally collected for the PI-CAI study~\citep{Saha2024}. The PI-CAI training set of 9,107 examinations is used for training; its 1,000-examination test set serves as an internal test cohort.}
  \item \rev{\textbf{Secondary training and internal testing cohort:} 10,264 examinations (10,149 patients) acquired from 2012 to 2024 in 22 cities in 13 countries in Europe, Asia, North America, and Africa, collected through the EU Horizon projects ProCAncer-I and COMFORT and eight independent centres. Of these, 9,264 examinations are used for training and 1,000 are held out as a second internal test cohort.}
\end{itemize}

\noindent \rev{The SCARLET-1 study tested PI-CAI-2B externally on the PRIME~\citep{Ng2025}, STHLM3-MRI~\citep{Eklund2021}, and IP1-PROSTAGRAM~\citep{EldredEvans2021} cohorts. The figures above are those of the study protocol; the published report gives the final composition of 22,445 examinations from 21,252 men~\citep{SahaSCARLET2026}. The internal test cohorts of the SCARLET-1 study are distinct from the UCLH internal test set of the present study.}

\rev{The weights of PI-CAI-2B are fixed throughout this study. UCLH, ReIMAGINE, and BiopsyUS are} not \rev{named among its development cohorts, and PRIME, STHLM3-MRI, and IP1-PROSTAGRAM were} used \rev{in its development study for external testing only. Case-level lists of the development cohorts are not available, so patient-level overlap with the cohorts of this study cannot be excluded; this is reported as a limitation.}

\smallskip
The UCLH cohort is then split into \rev{an adaptation} set \rev{(10}\%\rev{), a validation set (10\%),} and a held-out \rev{internal} test set \rev{(80}\%), stratified by csPCa status and grouped by patient, so that all cases from one patient fall in the same split. Grouping is necessary because men on active surveillance contribute serial studies that are highly correlated; splitting at case level would place near-duplicate images on both sides of the partition. The split is performed once using a fixed random seed. The \rev{adaptation} set trains the clinical \rev{modulation modules, with all PI-CAI-2B weights frozen; the validation set is used for model selection, for Platt recalibration,} and \rev{to fix decision thresholds;} and the test set is not used at any point during development.

% \begin{notebox}
% \textbf{No data leakage safeguard:} Normalisation statistics (mean, standard deviation) for clinical features are computed from the fine-tuning split only and applied to the test data.
% \end{notebox}

\subsection{Model Architecture}

For consistency, model terminology used throughout the study is defined in Table~\ref{tab:model_terminology}.

\begin{table}[htbp]
\centering
\caption{Model terminology.}
\label{tab:model_terminology}
\begin{tabular}{@{}L{3.5cm}L{9.0cm}@{}}
\toprule
\textbf{Term} & \textbf{Definition} \\
\midrule

Image-only model &
Model using bpMRI predictors alone (T2W, high b-value DWI, and \rev{ADC): PI-CAI-2B in its image-only configuration, with fixed weights (Section~\ref{sec:image-model}).} \\

\addlinespace

Context-only model &
Model using clinical predictors without \rev{the image encoder. Group~III predictors, although computed from the MRI, enter as scalar features and are included.} If not specified, this refers to the full clinical predictor set. \\

\addlinespace

Image-context model &
Any model combining bpMRI with one or more clinical predictors. \\

\addlinespace

Full clinical predictor set &
All clinical predictors included in Table~\ref{tab:clinical_vars}. \\

\addlinespace

Full image-context model &
Image-context model combining bpMRI with the full clinical predictor set. \\

\bottomrule
\end{tabular}
\end{table}

\subsubsection{Image Encoder}
\label{sec: image encoder}

\rev{The image encoder is the MedNeXt-3XL} network of \rev{PI-CAI-2B, fold~0 (Section~\ref{sec:image-model}), which} encodes the three-channel MRI input (T2W, \rev{high $b$-value} DWI, \rev{and ADC) prepared} as \rev{in Section~\ref{sec:imaging}. All of its weights are frozen} and are not updated \rev{with} UCLH data. For image-context models, \rev{only the} clinical \rev{modulation modules of Section~\ref{sec: clinical encoder}} are \rev{trained.}

\subsubsection{Clinical Feature Encoder}
\label{sec: clinical encoder}

\rev{Clinical context enters the frozen image model by feature-wise linear modulation (FiLM)~\citep{Perez2018}, applied per variable.} Each clinical \rev{variable $j$ has its own small MLP, which maps its value $c_j$ to a channel-wise scale $a_j(c_j)$ and shift $b_j(c_j)$. A signal $h$ of the image model} is \rev{modulated as}
\[\revon
  h' = h + \sum_{j} m_j \bigl[ a_j(c_j) \, h + b_j(c_j) \bigr],
\]
\rev{where $m_j$ indicates whether variable $j$ is recorded (Section~\ref{sec: missing data}). The output layer of every MLP is initialised to zero, so modulation starts from the image-only model.}

\subsubsection{Combining Strategies}
\label{sec: combining strategy}

Three image-context combining strategies will be \rev{compared. They use} the \rev{same per-variable modulation and differ only in} the \rev{signal $h$ that it modulates (Table~\ref{tab:fusion}). Beyond its technical interest, the comparison may indicate how clinical context interacts with imaging findings, and so generate hypotheses for further study.}

\begin{table}[htbp]
\centering
\caption{Image-context combining strategies.}
\label{tab:fusion}
\begin{tabular}{@{}L{2.8cm}L{9.7cm}@{}}
\toprule
\textbf{Strategy} & \rev{\textbf{Modulated signal $h$}} \\
\midrule
Early combining & \rev{Output of the first encoder block:} image feature maps \rev{at full spatial resolution.} \\
Mid combining & \rev{Output of} the \rev{encoder bottleneck: the most compressed image representation.} \\
Late combining & \rev{Case-level score of the image model, on the logit scale.} \\
\bottomrule
\end{tabular}
\end{table}

These strategies represent different ways in which clinical information may contribute to diagnostic reasoning. Early combining \rev{lets} clinical context \rev{change how} imaging \rev{features are formed.} Mid combining \rev{lets it adjust the summary} of \rev{the image.} Late combining \rev{changes only the final} risk \rev{estimate, as \emph{clinical risk calculators or nomogram-based decision support} adjust risk} at the decision \rev{level.}

\subsubsection{\rev{Training and Calibration}}
\rev{\label{sec:training}}

\rev{Only the modulation modules are trained, on the adaptation set, with} a \rev{binary cross-entropy loss. For early and mid combining, the case score of the full inference pipeline cannot be differentiated, so the loss} is \rev{computed on a proxy case score taken} from \rev{the top-64 lesion logits. Model selection uses} the \rev{validation set. Every model, including} the \rev{image-only model, is then recalibrated by Platt scaling on} the \rev{validation set, and this calibration} is \rev{fixed before} the \rev{test set} is \rev{evaluated.}

\subsection{Performance Metrics}

The primary objective is a comparison, so the primary estimand is the \emph{difference} between
the full image-context model and the image-only model on the held-out UCLH test set, reported
with a 95\% confidence interval rather than as two separate point estimates. Because both models
are applied to the same cases, all comparisons are paired: differences in AUC-ROC are tested
using DeLong's method~\citep{DeLong1988}, and differences in every other metric are obtained by
paired bootstrap resampling of cases with 2{,}000 replicates. The context-only model is reported
alongside as a reference for the performance attainable without imaging, and is not the subject
of a formal comparison.

A model may raise discrimination while worsening calibration, or improve average discrimination without improving performance at the operating point where a biopsy decision is actually made, and these are different failures with different
clinical consequences. The metrics in Table~\ref{tab:metrics} are therefore prespecified as a
panel spanning four distinct questions - overall discrimination, performance at a clinically
relevant operating point, calibration, and net clinical benefit - and a claim that clinical
context improves detection requires improvement in discrimination without degradation in
calibration.

\emph{Every metric in the panel is reported for every model, in both directions, irrespective of
whether the result favours the multimodal model.} No metric is selected, added, or dropped after
the test set has been examined. This constraint is what allows a panel of metrics to be more
informative than a single one.

Threshold-dependent metrics require a prespecified operating point. Sensitivity, specificity,
PPV, and NPV are reported at the threshold giving 95\% sensitivity in the \rev{validation} set,
reflecting the clinical priority of not missing csPCa, and additionally at the threshold giving
90\% sensitivity. Thresholds are fixed on the \rev{validation} set and applied unchanged to the test
set.

Only the primary comparison is confirmatory. All secondary objectives, including the ablation
analyses, are estimation exercises reported with confidence intervals\rev{, both without and with
adjustment for multiple comparisons}; given the number of comparisons involved, individual ablation
results are interpreted as ranking evidence.

\begin{table}[htbp]
\centering
\caption{Prespecified panel of performance metrics. All are reported for every model.}
\label{tab:metrics}
\begin{tabular}{@{}L{3.3cm}L{5.6cm}L{4.4cm}@{}}
\toprule
\textbf{Metric} & \textbf{Definition} & \textbf{Question it answers} \\
\midrule
AUC-ROC & Area under the receiver operating characteristic curve & Overall discrimination, independent of threshold \\
\addlinespace
AUPRC & Area under the precision-recall curve & Discrimination in the positive class, which ROC can flatter \\
\addlinespace
Sensitivity, specificity & At the prespecified operating points (95\% and 90\% sensitivity) & Performance where the biopsy decision is made \\
\addlinespace
PPV, NPV & Predictive values at the same operating points & How a prediction should change management \\
\addlinespace
Calibration & Calibration plot, calibration slope and intercept, Brier score & Whether predicted risks can be believed as probabilities \\
\addlinespace
Net benefit & Decision curve analysis across clinically plausible threshold probabilities & Whether adding clinical context is worth acting on \\
\bottomrule
\end{tabular}
\end{table}

\subsection{Image-only Model Evaluation}
\label{sec:image-only-eval}

The image-only model is evaluated with frozen weights, preprocessing, and inference settings throughout; no evaluation cohort, internal or external, is used to update the model or its decision thresholds. All evaluations use the case-level score \rev{of} the \rev{model after Platt recalibration on the validation set (Section~\ref{sec:training}),} together with the metrics in Table~\ref{tab:metrics}.

\subsection{\rev{External Testing}}
\rev{\label{sec:external}}

\rev{The} image-only model \rev{and} the \rev{image-context} models will be evaluated on \rev{five external cohorts (Table~\ref{tab:external_testing}). PRIME, ReIMAGINE, and BiopsyUS come from diagnostic settings; STHLM3-MRI and IP1-PROSTAGRAM come from population-based screening. All weights and decision thresholds are fixed after development on UCLH data, and no external cohort is used to update them.}

\rev{The cohorts differ in which clinical predictors were recorded. A predictor that a cohort did not record is treated as missing and contributes nothing to} the \rev{image-context models (Section}~\rev{\ref{sec: missing data}). Each cohort therefore tests} the \rev{models under its own pattern of available clinical information. Only predictors whose definition} and \rev{timing match those of Table}~\rev{\ref{tab:clinical_vars} are mapped; other fields are treated as missing.}

Eligible cases \rev{have} T2W, ADC, and high \rev{$b$-value} DWI and an outcome \rev{that can be mapped} to ISUP grade group $\geq$2. Performance is reported separately for each \rev{cohort, together with the number of eligible cases, the number of csPCa-positive cases, and the completeness of each clinical predictor. The two screening cohorts} are \rev{included subject to confirmation of local data access; if either is unavailable, this is reported as a deviation from the protocol.}

\begin{table}[htbp]
\centering
\caption{\rev{External testing cohorts. Clinical predictors are those recorded in each cohort, before mapping to the definitions of Table~\ref{tab:clinical_vars}.}}
\label{tab:external_testing}
\small
\begin{tabular}{@{}L{3.0cm}L{4.2cm}L{3.5cm}L{3.5cm}@{}}
\toprule
\textbf{Cohort} & \rev{\textbf{Setting and population}} & \rev{\textbf{Clinical predictors}} & \rev{\textbf{Outcome}} \\
\midrule
\rev{PRIME\newline\citep{Ng2025}} & \rev{Primary diagnosis. International multicentre randomised trial comparing biparametric with multiparametric MRI. External test cohort of the SCARLET-1 study.} & \rev{Age, PSA, prostate volume, PSA density, ethnicity, DRE, family history of prostate cancer, comorbidities, obesity} & \rev{Biopsy grade group; unbiopsied men are labelled only as the trial reference standard allows} \\
\addlinespace
\rev{ReIMAGINE\newline\citep{Marsden2022}} & \rev{Primary diagnosis. UK multicentre prospective cohort of men referred for MRI with suspected prostate cancer and a suspicious MRI (score $\geq$3).} & \rev{PSA, prostate volume} from the \rev{MRI report; PSA density derived} & \rev{Biopsy Gleason grade; the overall} or \rev{maximal grade rule is fixed before analysis} \\
\addlinespace
\rev{BiopsyUS\newline\citep{BiopsyUSRegistry,Min2025}} & \rev{Diagnosis. US registry of men undergoing MRI/ultrasound-guided biopsy (NCT02381990); 1,898 patients, 1,976 studies.} & \rev{Age, derived} from \rev{dates; prostate volume from segmentation} & \rev{Core-level ISUP grade group; case label from the highest grade} \\
\addlinespace
\rev{STHLM3-MRI\newline\citep{Eklund2021}} & \rev{Population-based screening, Stockholm, Sweden. External test cohort of the SCARLET-1 study.} & \rev{Age, PSA, prostate volume} & \rev{Grade group $\geq$2 by the trial reference standard} \\
\addlinespace
\rev{IP1-PROSTAGRAM\newline\citep{EldredEvans2021}} & \rev{Population-based screening, London, UK. External test cohort of the SCARLET-1 study.} & \rev{Age, PSA, prostate volume, ethnicity} & \rev{Grade group $\geq$2 by the trial reference standard; few positive} cases \rev{limit precision} \\
\bottomrule
\end{tabular}
\end{table}

\subsection{Ablation Studies}

To quantify the contribution of individual clinical predictors to multimodal prediction, add-one-in (AOI) and leave-one-out (LOO) analyses will be performed separately for each combining strategy:

\begin{itemize}[leftmargin=2em]
  \item \textbf{Add-one-in:} The image-only model serves as the reference. Each clinical predictor is added individually to bpMRI using the specified combining strategy. Its effect is assessed by the change in model performance relative to the image-only model.
  \item \textbf{Leave-one-out:} The full image-context model serves as the reference. Each clinical predictor is removed individually while all remaining clinical predictors are retained. Its effect is assessed by the change in model performance relative to the full image-context model.
\end{itemize}

Each ablation experiment constitutes an independent full training run to avoid masking effects.

\subsection{Determinants of Variable Contribution}
\label{sec:determinants}

The incremental value of each clinical predictor, estimated by the add-one-in analysis, will be related to three characteristics of that predictor: its source (structured extract, derivation, automated \rev{image analysis,} or extraction from free text), its missingness rate in the \rev{adaptation} set, and, for free-text-derived variables, its validated extraction reliability (Table~\ref{tab:supp_agreement}). The association will be summarised descriptively, principally as incremental performance plotted against missingness rate with points coloured by source; given the small number of predictors, it will not be formally tested.

This analysis is exploratory. It asks whether variables that are harder to obtain reliably also contribute less, which bears on which clinical information is worth the cost of capturing in a future prospective system.

\subsection{Interpretability Analysis}

SHAP (SHapley Additive exPlanations)~\citep{Lundberg2017} values will be computed on the test set to quantify the contribution of each clinical feature to individual predictions. Summary plots (beeswarm and bar plots) will be presented for each combining strategy.

\subsection{Subgroup Analyses}
\label{sec: subgroup analysis}
\rev{Subgroup analyses ask whether the contribution of clinical context differs between groups of patients, scans, or lesions. The subgroups and their cut-off points are fixed before the test set is analysed. Clinical thresholds are used wherever an established one exists; where none exists, the cut-off points are the tertiles of the adaptation set.}

\begin{itemize}[leftmargin=2em]
  \item \rev{\textbf{Patient characteristics:}}
  \begin{itemize}
    \item \rev{Age:} $<$60, \rev{60--69,} $\geq$70 years.
    \item \rev{Ethnicity:} White, Black, Asian, Other/Not stated. \rev{The last subgroup combines the Mixed, Other, and Unknown categories of Table~\ref{tab:clinical_vars}.}
    \item \rev{BMI: $<$25, 25--29.9, $\geq$30~kg/m$^2$.}
    \item \rev{Smoking status: never, former, current.}
    \item \rev{Diabetes: present, absent.}
  \end{itemize}
  \item \rev{\textbf{Inherited risk:} family history of prostate cancer (present, absent); family history of breast or ovarian cancer (present, absent).}
  \item \rev{\textbf{PSA and examination:}}
  \begin{itemize}
    \item \rev{PSA level:} $<$4, \rev{4--10, 10--20,} $>$20\rev{~}ng/mL.
    \item \rev{PSA density: $<$0.10, 0.10--0.19, $\geq$0.20~ng/mL/mL.}
    \item \rev{DRE finding: normal, abnormal.}
  \end{itemize}
  \item \rev{\textbf{Treatments that alter PSA or the gland:} 5-ARI use at MRI (yes, no); prior BPH intervention (yes, no).}
  \item \rev{\textbf{Radiologist-derived context:}}
  \begin{itemize}
    \item \rev{Radiologist score: 1--2,} 3, 4, 5.
    \item \rev{Number of lesions with radiologist score $\geq$3: 0, 1, $\geq$2.}
    \item \rev{Index lesion zone: peripheral zone, transition zone, other or multiple zones. Restricted to cases with a reported lesion.}
  \end{itemize}
  \item \rev{\textbf{Image-derived measurements:}}
  \begin{itemize}
    \item \rev{Prostate volume: $<$40, 40--79, $\geq$80~mL.}
    \item \rev{PPAT measurements, each divided at tertiles: PPAT volume, PPAT / prostate volume ratio, PPAT thickness, PPAT fat fraction, and retropubic gap.}
  \end{itemize}
  \item \rev{\textbf{Image-only model score:} low-score (below the median score of the image-only model in the adaptation set) and high-score (at or above that median).}
  \item \rev{\textbf{Acquisition:} field strength (1.5T, 3T); high $b$-value series acquired or computed (Section~\ref{sec:imaging}).}
\end{itemize}

\rev{For a variable that can be missing, cases in which it was not recorded form a separate \emph{not recorded} subgroup. This subgroup is reported because missingness itself may carry information. Subgroups are defined by the recorded values of the variables, whether or not those variables are inputs to the model under evaluation.}

\rev{Subgrouping by image-only model score uses the output of one of the models being compared. Selecting on this score restricts its range within each subgroup, so the discrimination of the image-only model within a subgroup is expected to be lower than that in the whole cohort, and the difference between models is interpreted accordingly.}

\rev{Within each subgroup, every model is evaluated with the metrics of Table~\ref{tab:metrics}. The difference between the full image-context model and the image-only model is reported with a 95\% confidence interval, estimated as in the primary comparison. Whether this difference varies across the levels of a subgrouping variable is assessed by an interaction estimate obtained by paired bootstrap resampling.}

\rev{A subgroup with fewer than 30 csPCa-positive or 30 csPCa-negative cases is merged with the adjacent level, or is reported descriptively where no adjacent level exists. All subgroup analyses are exploratory. They are reported in full, without adjustment for multiple comparisons, and are not used to support claims about the primary objective.}

\subsection{Sensitivity Analyses}
\label{sec:sensitivity}

\begin{enumerate}[label=\arabic*., leftmargin=2em]
  \item \textbf{Verification bias correction:} Inverse probability weighting (IPW) to account for differential biopsy rates between radiologist score categories.
  \item \textbf{Complete case analysis:} Repeating primary analyses restricted to patients with $<$20\% missing clinical variables.
  \item \textbf{Alternative outcome definitions:} Using any-grade prostate cancer (ISUP $\geq$1) as an alternative outcome.
  \item \textbf{Pathology linkage interval:} Repeating the labelling procedure with intervals of 6 and 12~months in place of 3~months, to quantify the effect of the linkage window on the proportion of cases labelled by histopathology and on model performance. Relabelling does not require the image encoder to be retrained.
  \item \rev{\textbf{PPAT band width:} Repeating the analyses with PPAT defined at 10~mm and 20~mm from the prostate surface in place of 15~mm.}
  \item \rev{\textbf{Gland truncation:} Repeating the analyses after excluding cases in which the prostate mask reaches the boundary of the field of view (Section~\ref{sec:image-derived}).}
\end{enumerate}

\subsection{Software}

All analyses will be performed in Python 3.10+ using PyTorch (deep learning), scikit-learn (classical ML comparisons), and SHAP (interpretability). Statistical comparisons between AUCs will use DeLong's test~\citep{DeLong1988}. Code and trained model weights will be made available upon publication.

% ==========================================================
\section{Bias Assessment}
\label{sec:bias}

This study is subject to several domain-specific biases that are intrinsic to the prostate cancer diagnostic pathway and the retrospective study design. These are discussed below alongside general methodological biases.

\subsection{PSA Referral Bias}

The cohort is conditioned on referral for prostate MRI, which in clinical practice is triggered primarily by elevated or rising PSA. This creates a \emph{PSA referral bias}: patients entering the cohort have, by definition, already been flagged by PSA testing, and the distribution of PSA values in this referred population does not reflect the general male population. Consequently, the predictive value of PSA within this cohort may be attenuated compared with a screening population (``PSA has already done its job'' by the time MRI is performed). Models trained on this cohort may overestimate the marginal contribution of imaging relative to PSA, and performance metrics may not generalise to settings where PSA-based pre-selection has not occurred (e.g., opportunistic MRI screening).

\subsection{Patient Selection Bias}

UCLH is a high-volume tertiary referral centre with particular expertise in prostate MRI. The patient population is therefore enriched for complex cases, higher radiologist scores, and a higher baseline prevalence of csPCa (34\%) compared with community or screening settings. Additionally, the ethnic and socioeconomic composition of the UCLH catchment area (central London) may not be representative of the wider UK or international populations. These factors limit the direct generalisability of model performance estimates. External validation on datasets from community hospitals and diverse populations will be essential before clinical deployment.

\subsection{Classification Task Bias}

The outcome definition (csPCa as ISUP $\geq$2 on biopsy) introduces a \emph{classification task bias} with several components:

\begin{itemize}[leftmargin=2em]
  \item \textbf{Verification bias:} Not all patients undergo biopsy; those with a radiologist score of 1--2 are less likely to be biopsied, leading to under-ascertainment of csPCa in low-risk groups. The magnitude of this bias 
  will be assessed using inverse probability weighting (IPW), as specified in Sec.~\ref{sec:sensitivity}, with weighted and unweighted model performance compared.
  
  \item \textbf{Biopsy sampling error:}  Targeted and systematic biopsies may miss significant cancer, with studies reporting approximately 5-16\% of csPCa being missed when systematic biopsy is omitted and around 9\% misclassification of clinically significant disease with MRI-targeted biopsy alone (\cite{Ahdoot2020, Lee2022}).
  
  \item \textbf{Gleason grading variability:} Inter-observer agreement for Gleason grading, particularly at the 3+3 vs.\ 3+4 boundary, is imperfect. The binary threshold (ISUP $\geq$2) is clinically standard but introduces classification noise at the decision boundary. A UK reproducibility study reported 78\% agreement with the consensus Gleason score and an overall inter-observer $\kappa$ of 0.54, with lower agreement for Gleason score 7 ($\kappa=0.44$) (\cite{Melia2006}).

  \item \textbf{Temporal label evolution:}
   Imaging interpretation, biopsy practice, and pathology reporting may evolve over the study period, resulting in temporal heterogeneity in outcome ascertainment.
  % For example, PI-RADS v2.1, introduced in 2019, refined several v2 assessment criteria, including the evaluation of
  % transition-zone nodules and the use of DWI to upgrade atypical transition-zone lesions.
  % [commented out: UCLH reports on the Likert scale, so PI-RADS version changes do not act on these labels]
  % Pathology reporting practices and the proportion of MRI-targeted vs.\ systematic biopsies may have shifted over the five-year study period, introducing temporal heterogeneity in outcome ascertainment.
\end{itemize}

\subsection{Other Methodological Biases}

Several methodological biases may influence the estimated contribution of clinical context and
the interpretation of model performance. Their anticipated impact on this study and corresponding
mitigation strategies are summarised below.

\begin{table}[htbp]
\centering
\caption{Additional sources of bias and mitigation strategies.}
\label{tab:bias}
\begin{tabular}{@{}L{2.5cm}L{4.3cm}L{1.8cm}L{6.0cm}@{}}
\toprule
\textbf{Bias Type} & \textbf{Risk} & \textbf{Expected Impact} & \textbf{Mitigation and Remaining Issues} \\
\midrule

Information bias &
Variable MRI quality across scanners, protocols, and time; inter-reader variability in the radiologist score, and possible variation in the scale used &
Low &
Standardised preprocessing; as a pioneering protocol, remaining scanner-, protocol-, and reader-related effects will be further assessed through external validation  \\

\addlinespace

\rev{Measurement bias (image-derived features)} &
\rev{Adipose tissue classified by signal intensity rather than by a fat-specific sequence; seminal vesicle fluid may be counted as fat near the gland base; volumes underestimated where the gland exceeds the field of view} &
\rev{Moderate} &
\rev{Quality control and segmentation validation (Sec.~\ref{sec:image-derived}); sensitivity analyses on band width and gland truncation (Sec.~\ref{sec:sensitivity}); thresholds are cohort-specific and not transferable to other protocols} \\

\addlinespace

Confounding &
Unmeasured confounders (e.g., \rev{prior negative biopsy}) &
Low &
Adjustment using available covariates and sensitivity analyses; residual confounding remains an open research question \\

\addlinespace

Missing data bias &
Differential missingness (8--78\% across variables); LLM-extracted variables may have systematic extraction failures &
High &
Missing-data handling is variable-dependent and follows the procedures described in Sec.~\ref{sec: missing data} and~\ref{sec:llm-extraction}, including missingness-aware modelling, sensitivity analyses, and validation of LLM-derived variables against manual review \\

\addlinespace

Algorithmic bias &
Performance disparity across demographic subgroups &
Low &
Comparative evaluation across combining strategies (Sec.~\ref{sec: combining strategy}), together with subgroup analyses (Sec.~\ref{sec: subgroup analysis}), will identify potential algorithm-dependent disparities and inform future mitigation and external validation \\

\bottomrule
\end{tabular}
\end{table}

\section{Ethical Considerations}
\label{sec:ethics}

\subsection{Ethical Approval}

This study has been approved under the UCLH Data Access Process for Research (DAP-R), which provides a generic favourable ethical opinion (IRAS No.\ 299136). Individual patient consent is not required as this is a retrospective analysis of routinely collected, anonymised data. The study sponsor is UCLH.

\subsection{Data Protection and Privacy}

\begin{itemize}[leftmargin=2em]
  \item All data will be pseudonymised at extraction via the OMOP ES pipeline and stored in the UCL Data Safe Haven or UCLH Azure landing zones with RBAC.
  \item Patients registered on the National Data Opt-Out will be automatically excluded.
  \item DICOM tags will be anonymised prior to image conversion and analysis.
  \item Statistical disclosure control techniques will be applied to prevent re-identification through unique attribute constellations.
  \item The study complies with the UK General Data Protection Regulation (UK GDPR) and the Data Protection Act 2018. The lawful basis for processing is Article 6(1)(e) (public task) and Article 9(2)(j) (scientific research).
\end{itemize}

% \subsection{Patient and Public Involvement}

% A dedicated prostate cancer PPI (Patient and Public Involvement) group, established under an NIHR Research Professorship award, has been consulted on the use of AI in prostate cancer diagnosis. The group endorsed the development of AI tools to assist radiologists and expressed support for the staged development approach. Ongoing engagement includes face-to-face meetings and a public open day.

% ==========================================================
\section{References}
\label{sec:references}

\begingroup
\renewcommand{\section}[2]{}  % suppress bibliography heading

\endgroup

% ==========================================================
\appendix
\renewcommand{\thetable}{S\arabic{table}}
\renewcommand{\theHtable}{S\arabic{table}}  % unique hyperref anchors
\setcounter{table}{0}

\section{Supplementary Material}
\label{sec:supplementary}

\subsection{Radiology Report Keyword Rules}
\label{sec:supp-keywords}

The keyword rules in Table~\ref{tab:supp_keywords} apply only at step~4 of the outcome labelling procedure (Section~\ref{sec:variables}), that is, only to cases for which no histopathology, no recorded lesion count, and no assessment score is available. They are a last-resort rule set applied to a minority of cases, and they are deliberately conservative: a case is labelled csPCa-negative only when the report contains an explicit statement of absence, and any expression of diagnostic uncertainty prevents the label from being assigned.

Rules are applied to the Conclusion or Impression section of the report where one can be identified, and to the full report text otherwise, since the concluding statement carries the reporting radiologist's overall assessment. Matching is case-insensitive and follows whitespace and punctuation normalisation.

Exclusion patterns (E1--E8) are evaluated before inclusion patterns (N1--N9). A case matching any exclusion pattern is not labelled at this step and is excluded from the cohort, because step~4 is the final opportunity for ascertainment. Negation is resolved using a scope-limited negation detection method of the NegEx family rather than by literal string matching, so that a negation term applies only to findings within the same clause; this prevents a phrase such as ``no extracapsular extension'' from negating a suspicious lesion described in an adjacent sentence.

The rule set will be frozen, with a version identifier, before it is applied to the study cohort. Any subsequent amendment will be recorded with its date, rationale, and the number of cases whose label changes as a result.

\begin{longtable}{@{}L{1.1cm}L{7.3cm}L{5.2cm}@{}}
\caption{Radiology report keyword rules applied at step~4 of outcome labelling. Patterns are shown in a simplified regular-expression notation; alternation is denoted by a vertical bar and optional elements by a question mark.}
\label{tab:supp_keywords} \\
\toprule
\textbf{ID} & \textbf{Pattern (case-insensitive)} & \textbf{Rationale} \\
\midrule
\endfirsthead
\toprule
\textbf{ID} & \textbf{Pattern (case-insensitive)} & \textbf{Rationale} \\
\midrule
\endhead
\bottomrule
\endfoot

\multicolumn{3}{@{}L{13.6cm}@{}}{\textbf{Exclusion patterns} (evaluated first; a match prevents any label being assigned)} \\
\addlinespace[2pt]

E1 & \texttt{suspicio(us|n)}, \texttt{concerning} (unnegated) & Any surviving expression of suspicion is incompatible with an explicitly negative report \\
E2 & \texttt{equivocal}, \texttt{indeterminate}, \texttt{borderline} & Diagnostic uncertainty; the case should be resolved by biopsy rather than by rule \\
E3 & \texttt{cannot (be )?exclude(d)?}, \texttt{not excluded} & Explicit non-exclusion of malignancy \\
E4 & \texttt{recommend(ed)? biopsy}, \texttt{for biopsy}, \texttt{proceed to biopsy} & Onward action implying residual suspicion \\
E5 & \texttt{extracapsular extension}, \texttt{seminal vesicle invasion}, \texttt{ECE}, \texttt{SVI} & Staging features of established disease \\
E6 & \texttt{metasta(sis|\allowbreak ses|\allowbreak tic)}, \texttt{lymphadenopathy}, \texttt{bone lesion} & Disease beyond the prostate \\
E7 & \texttt{progression}, \texttt{interval increase}, \texttt{enlarging} & Change implying active disease, relevant to active surveillance studies \\
E8 & \texttt{(probabl(e|y)|\allowbreak likely|\allowbreak favour(s|ed)?)} within the same clause as a malignancy term & Hedged positive assertion \\

\addlinespace
\multicolumn{3}{@{}L{13.6cm}@{}}{\textbf{Inclusion patterns} (a match in the absence of any exclusion match labels the case csPCa-negative)} \\
\addlinespace[2pt]

N1 & \texttt{no (suspicious|\allowbreak significant|\allowbreak concerning) (lesion|\allowbreak abnormality|\allowbreak focus|\allowbreak area)} & Explicit absence of a suspicious finding \\
N2 & \texttt{no evidence of (clinically significant )?\allowbreak (prostate )?(cancer|\allowbreak malignancy|\allowbreak tumour)} & Explicit absence of malignancy \\
N3 & \texttt{screen(ing)? negative} & Local reporting convention for a negative study \\
N4 & \texttt{normal (prostate )?\allowbreak (MRI|study|\allowbreak appearances)} & Global statement of normality \\
N5 & \texttt{unremarkable (prostate|\allowbreak appearances|\allowbreak study)} & Global statement of normality \\
N6 & \texttt{no (focal )?(lesion|\allowbreak abnormality) (identified|\allowbreak seen|\allowbreak detected|\allowbreak demonstrated|\allowbreak visualised)} & Explicit absence of a focal finding \\
N7 & \texttt{benign (appearances|features|changes)} not accompanied by any E-pattern & Positive assertion of benignity \\
N8 & \texttt{no target(able)? lesion} & Explicit absence of a biopsy target \\
N9 & \texttt{no MRI evidence of (significant )?disease} & Explicit absence of disease on MRI \\

\end{longtable}

\subsection{LLM Extraction Schema and Prompt}
\label{sec:supp-schema}

Table~\ref{tab:supp_schema} gives the extraction schema submitted to the model, one row per target variable. The schema is enforced twice: it is stated in the prompt, and parsed outputs are validated against it before any value enters the analysis dataset (Section~\ref{sec:llm-extraction}). The system prompt reproduced below is applied unchanged to every patient; the per-variable instructions in Table~\ref{tab:supp_schema} are appended to it as the JSON schema description.

\begin{small}
\begin{verbatim}
You extract structured clinical data from prostate cancer records.

Rules:
1. Return a single JSON object matching the schema. No prose, no markdown.
2. Return null for any field not stated in the source text. Never infer,
   estimate, or carry a value forward from another field.
3. Return false only where the text explicitly records absence. Absence of
   mention is null, not false.
4. Where a value appears more than once with different values, apply the
   aggregation rule given in the field description (for example, maximum
   for radiologist scores, highest for Gleason score).
5. Do not infer a diagnosis from medication alone.
6. Quote the source verbatim in the accompanying _evidence field for every
   non-null value. If you cannot quote it, return null.
7. Report dates in ISO 8601 (YYYY-MM-DD). Return null for partial dates.
\end{verbatim}
\end{small}

\noindent Rule~6 requires the model to return a supporting quotation alongside each extracted value. Values whose quotation cannot be located in the source document by exact string match are set to null during post-processing. This converts an unverifiable assertion into a detectable failure and provides the source linkage described in Section~\ref{sec:llm-extraction}.

\begin{longtable}{@{}L{2.5cm}L{3.2cm}L{8.0cm}@{}}
\caption{Per-variable extraction schema, covering only those variables sourced from free text (Table~\ref{tab:clinical_vars}, source \emph{LLM}) together with the report fields used for outcome labelling. Every field additionally accepts \texttt{null}, and every field has a paired \texttt{\_evidence} field holding the supporting quotation.}
\label{tab:supp_schema} \\
\toprule
\textbf{Variable} & \textbf{Permitted values} & \textbf{Extraction instruction} \\
\midrule
\endfirsthead
\toprule
\textbf{Variable} & \textbf{Permitted values} & \textbf{Extraction instruction} \\
\midrule
\endhead
\bottomrule
\endfoot

\multicolumn{3}{@{}L{13.7cm}@{}}{\textbf{From radiology reports}} \\
\addlinespace[2pt]

Radiologist score & Integer 1--5 & Return the highest level-of-suspicion score stated for any lesion, whether reported as a PI-RADS category or as a Likert score. Where several lesions are scored, return the maximum. Record which scale was used in the paired \texttt{score\_scale} field. Do not infer a score from a narrative description of suspicion. \\
Lesion count & Integer $\geq$0 & Count lesions explicitly enumerated or individually described. Where the report states that no lesion is present, return 0. Where lesions are referred to collectively without enumeration, return null. \\

\addlinespace
\multicolumn{3}{@{}L{13.7cm}@{}}{\textbf{From histopathology reports}} \\
\addlinespace[2pt]

Gleason score & String, \texttt{a+b} & Return the highest Gleason score across all cores in the specimen. Where primary and secondary patterns are reported separately, combine them in the stated order. \\
ISUP grade group & Integer 1--5 & Return the grade group as stated. Where only a Gleason score is given, return null; the mapping is applied during post-processing rather than by the model. \\

\addlinespace
\multicolumn{3}{@{}L{13.7cm}@{}}{\textbf{From clinical notes}} \\
\addlinespace[2pt]

Family history of prostate cancer & Boolean & Return true only where a blood relative with prostate cancer is recorded. Return false only where family history is explicitly documented as negative. Relatives by marriage do not count. \\
Family history of any cancer & Boolean & As above, for any malignancy in a blood relative. Where true, list the cancer types in the paired \texttt{family\_cancer\_types} array. \\
History of other cancers & Boolean & Return true where the patient has a personal history of a malignancy other than prostate cancer. Benign neoplasms and in-situ lesions do not count. Where true, list types in \texttt{other\_cancer\_types}. \\
Diabetes & Boolean & Return true where recorded as a current or past diagnosis, including type~1 and type~2. Do not infer from glucose-lowering medication alone. \\
Hypertension & Boolean & Return true where recorded as a diagnosis. Do not infer from antihypertensive medication or from isolated blood pressure readings. \\
Cardiovascular disease & Boolean & Return true for ischaemic heart disease, myocardial infarction, heart failure, atrial fibrillation, stroke, or peripheral arterial disease. Hypertension alone does not count. \\
COPD & Boolean & Return true where recorded as a diagnosis. Emphysema and chronic bronchitis count; asthma does not. \\
Asthma & Boolean & Return true where recorded as a diagnosis. \\
Chronic kidney disease & Boolean & Return true where recorded as a diagnosis or where a CKD stage is stated. Do not infer from a single creatinine or eGFR value. \\
Smoking status & Integer 0--2 & 0~= never smoker; 1~= former smoker; 2~= current smoker. Where the text records only ``smoker'' without qualification, return 2. \rev{The integer code is one-hot encoded during post-processing.} \\
Alcohol \rev{status} & Integer 0--6 & Ordinal band of reported weekly consumption in UK units: 0~= none; 1~= 1--7; 2~= 8--14; 3~= 15--21; 4~= 22--35; 5~= 36--50; 6~= $>$50. Where consumption is described qualitatively without a quantity, return null. \rev{The integer code is one-hot encoded during post-processing.} \\

\end{longtable}

\subsection{Duplicate Extraction and Adjudication}
\label{sec:supp-agreement}

Table~\ref{tab:supp_agreement} is the reporting template for the two checking procedures described in Section~\ref{sec:llm-extraction}: duplicate extraction, which quantifies the stability of the extraction process, and adjudicated re-review, which tests the reference standard against which the process is judged. Both are reported per variable, since reliability differs markedly between variables that are stated explicitly in structured-like text and those that must be inferred from narrative.

\begin{longtable}{@{}L{2.7cm}C{1.5cm}C{2.0cm}C{2.0cm}C{1.6cm}L{3.6cm}@{}}
\caption{Reporting template for duplicate extraction and adjudicated re-review. One row per variable.}
\label{tab:supp_agreement} \\
\toprule
\textbf{Variable} & \textbf{n} & \textbf{Same-model agreement} & \textbf{Between-model agreement} & \textbf{F$_1$ vs consensus} & \textbf{Discrepancies by cause} \\
\midrule
\endfirsthead
\toprule
\textbf{Variable} & \textbf{n} & \textbf{Same-model agreement} & \textbf{Between-model agreement} & \textbf{F$_1$ vs consensus} & \textbf{Discrepancies by cause} \\
\midrule
\endhead
\bottomrule
\endfoot

\multicolumn{6}{@{}L{13.4cm}@{}}{\emph{One row per variable in Table~\ref{tab:supp_schema}; completed on study execution.}} \\

\end{longtable}

\noindent Discrepancies are classified as source ambiguity, annotator error, or extraction error. The distinction matters because only extraction error is attributable to the pipeline and correctable by prompt revision; source ambiguity sets a ceiling on the accuracy any extraction method could achieve for that variable, and is reported as such rather than counted against the model.

\end{document}